\documentclass{nle}

\usepackage{theapa-modified}         
\usepackage{amsmath}                 
\usepackage{url}                     
\usepackage{float}                   

\title[Three Approaches to Recognizing Lexical Entailment]       
      {Experiments with Three Approaches to \\                   
       Recognizing Lexical Entailment}

\author[P. D. Turney and S. M. Mohammad]                         
       {P. D.\ns T\ls U\ls R\ls N\ls E\ls Y\ls and\ns            
        S. M.\ns M\ls O\ls H\ls A\ls M\ls M\ls A\ls D\\
        National Research Council Canada\\
        Ottawa, Ontario, Canada, K1A 0R6.\\
        \{peter.turney, saif.mohammad\}$@$nrc-cnrc.gc.ca}

\pagerange{\pageref{firstpage}--\pageref{lastpage}}

\pubyear{2014}

\begin{document}

\label{firstpage}

\maketitle

\begin{abstract}
Inference in natural language often involves recognizing lexical entailment (RLE);
that is, identifying whether one word entails another. For example, {\em buy} entails
{\em own}. Two general strategies for RLE have been proposed: One strategy is to
manually construct an asymmetric similarity measure for context vectors ({\em directional
similarity}) and another is to treat RLE as a problem of learning to recognize semantic
relations using supervised machine learning techniques ({\em relation classification}).
In this paper, we experiment with two recent state-of-the-art representatives of the two
general strategies. The first approach is an asymmetric similarity measure (an instance
of the {\em directional similarity} strategy), designed to capture the degree to which the
contexts of a word, $a$, form a subset of the contexts of another word, $b$. The second
approach (an instance of the {\em relation classification} strategy) represents a word pair,
$a\!:\!b$, with a feature vector that is the concatenation of the context vectors of $a$ and
$b$, and then applies supervised learning to a training set of labeled feature vectors.
Additionally, we introduce a third approach that is a new instance of the {\em relation
classification} strategy. The third approach represents a word pair, $a\!:\!b$, with a feature
vector in which the features are the differences in the similarities of $a$ and $b$ to a set
of reference words. All three approaches use vector space models (VSMs) of semantics, based
on word--context matrices. We perform an extensive evaluation of the three approaches using
three different datasets. The proposed new approach (similarity differences) performs
significantly better than the other two approaches on some datasets and there is no dataset
for which it is significantly worse. Along the way, we address some of the concerns raised
in past research, regarding the treatment of RLE as a problem of semantic relation classification,
and we suggest it is beneficial to make connections between the research in lexical entailment
and the research in semantic relation classification.
\end{abstract}

\section{Introduction}
\label{sec:intro}

Recognizing textual entailment (RTE) is a popular task in natural
language processing research, due to its relevance for text summarization,
information retrieval, information extraction, question answering, machine
translation, paraphrasing, and other applications \cite{androutsopoulos10}.
RTE involves pairs of sentences, such as the following \cite{dagan09}:

\begin{verse}

Text: {\em Norway's most famous painting, `The Scream' by Edvard Munch, was recovered
Saturday, almost three months after it was stolen from an Oslo museum.} \\*  
Hypothesis: {\em Edvard Munch painted `The Scream'.}

\end{verse}

\noindent The objective is to develop algorithms that can determine
whether the text sentence entails the hypothesis sentence. In RTE, the gold standard
for entailment is established by common sense, rather than formal logic. The text entails
the hypothesis if the meaning of the hypothesis can be inferred from the meaning of the
text, according to typical human interpretations of the text and the hypothesis \cite{dagan09}.
In the above example, the text entails the hypothesis. In recent years, the
RTE pairs have grown richer and more challenging. The text may now be a whole paragraph.

In many cases, to recognize when one {\em sentence} entails another, we must first be able to
recognize when one {\em word} entails another \cite{geffet05}. Consider this sentence pair
(our example):

\begin{verse}

Text: {\em George was bitten by a dog.} \\*  
Hypothesis: {\em George was attacked by an animal.}

\end{verse}

\noindent To recognize that the text entails the hypothesis, we must first recognize
that {\em bitten} entails {\em attacked} and {\em dog} entails {\em animal}.
This is the problem of recognizing lexical entailment (RLE). (We discuss the definition
of lexical entailment in Section~\ref{sec:defining}.)

Vector space models (VSMs) of semantics have been particularly useful for lexical
semantics \cite{turney10}, hence it is natural to apply them to RLE. In this paper,
we experiment with three VSM algorithms for lexical entailment. All three use
word--context matrices, in which a word corresponds to a row vector, called a
{\em context vector}. For a given word, the corresponding context vector represents
the distribution of the word over various contexts. The contexts consist of the words
that occur near the given word in a large corpus of text. These models are inspired by the
distributional hypothesis \cite{harris54,firth57}:

\begin{verse}

{\em Distributional hypothesis:} Words that occur in similar contexts tend to have
similar meanings.

\end{verse}

The first of the three algorithms, {\em balAPinc} (balanced average precision for distributional
inclusion), attempts to address the problem of RLE with an asymmetric similarity measure
for context vectors \cite{kotlerman10}. The idea is to design a measure that captures the
context inclusion hypothesis \cite{geffet05}:

\begin{verse}

{\em Context inclusion hypothesis:} If a word $a$ tends to occur in a subset of the
contexts in which a word $b$ occurs ($b$ contextually includes $a$), then $a$ (the
narrower term) tends to entail $b$ (the broader term).

\end{verse}

\noindent This is our paraphrase of what \citeA{geffet05} call the {\em distributional
inclusion hypothesis}. We prefer to call it {\em context inclusion}
rather than {\em distributional inclusion}, to be clear about what is included.
In the text--hypothesis example above, the narrower terms are {\em bitten}
and {\em dog} and the broader terms are {\em attacked} and {\em animal}.

The intent of balAPinc is to take the context vectors $\mathbf{a}$ and $\mathbf{b}$
for the words $a$ and $b$ and calculate a numerical score that measures the degree to which
$b$ contextually includes $a$. The context inclusion hypothesis is inspired by model theory
in formal logic \cite{hunter96}. Let $a$ and $b$ be assertions in formal logic. In model theory,
`$a \models b$' means $a$ entails $b$. If $a \models b$, then the set of models in which $a$ is
true is a subset of the set of models in which $b$ is true. That is, the models of $b$ include
the models of $a$.

The second and third algorithms approach the task of recognizing lexical entailment
by using techniques from research in semantic relation classification. Semantic relation
classification is the task of learning to recognize when a word pair is an instance of a given
semantic relation class \cite{rosario01,rosario02,nastase03,turney06b,girju07}.

An important subclass of lexical entailment is covered by the hyponymy--hypernymy semantic
relation. If a word pair $a\!:\!b$ is an instance of the hyponym--hypernym relation
({\em dog:animal}), then $a \models b$. There is a relatively large body of work on semantic
relation classification in general, with good results on the hyponym--hypernym relation in
particular \cite{hearst92,snow06}. Since semantic relation classification algorithms have
worked well for this important subclass of lexical entailment, it seems plausible that this
approach can be expanded to cover other subclasses of lexical entailment, and perhaps all
subclasses of lexical entailment. (We say more about this in Section~\ref{sec:relations-entailment}.)

The second of the three algorithms represents a word pair, $a\!:\!b$, with a feature vector that
is the concatenation of the context vector $\mathbf{a}$ for $a$ and the context vector $\mathbf{b}$
for $b$ \cite{baroni12}. For example, the concatenation of the two three-dimensional vectors
$\langle 1, 2, 3 \rangle$ and $\langle 4, 5, 6 \rangle$ is the six-dimensional vector
$\langle 1, 2, 3, 4, 5, 6 \rangle$. This algorithm was not given a name by \citeA{baroni12}.
For ease of reference, we will call it {\em ConVecs} (concatenated vectors).

ConVecs is based on the context combination hypothesis \cite{baroni12}:

\begin{verse}

{\em Context combination hypothesis:} The tendency of $a$ to entail $b$ is
correlated with some learnable function of the contexts in which $a$ occurs and
the contexts in which $b$ occurs; some conjunctions of contexts tend to indicate
entailment and others tend to indicate a lack of entailment.

\end{verse}

\noindent This hypothesis implies that the contexts of $a$ (the elements in
the context vector $\mathbf{a}$) and the contexts of $b$ (elements in $\mathbf{b}$)
are suitable features for a feature vector representation of the word pair $a\!:\!b$.
That is, if this hypothesis is correct, concatenated context vectors are an appropriate
representation of word pairs for supervised machine learning of lexical entailment.
This hypothesis was not explicitly stated by \citeA{baroni12} but it is implicit in
their approach.

In the semantic relation classification literature, vector concatentation (but not
necessarily with context vectors) is a common way to construct feature vectors for
supervised learning with word pairs \cite{rosario01,rosario02,nastase03}.
Context concatentation is a first-order feature vector representation of word pairs.
We call it {\em first-order} because the features are directly based on the
elements of the context vectors.

This paper introduces a new algorithm, {\em SimDiffs} (similarity differences), as the third of
the three algorithms we evaluate. SimDiffs uses a second-order feature vector representation of
$a\!:\!b$, in which the features are differences in the similarities of $a$ and $b$
to a set of reference words, $R$. The similarities are given by cosines of the first-order
context vectors for $a$, $b$, and the reference words, $r \in R$. (We use a set
of common words for $R$, as described in Section~\ref{subsec:simdiffs}. We do not
experiment with other choices for $R$.)

SimDiffs is dependent on the similarity differences hypothesis (introduced here):

\begin{verse}

{\em Similarity differences hypothesis}: The tendency of $a$ to entail $b$ is
correlated with some learnable function of the differences in their similarities,
${\rm sim}(a,r) - {\rm sim}(b,r)$, to a set of reference words, $r \in R$; some
differences tend to indicate entailment and others tend to indicate a lack of entailment.

\end{verse}

\noindent For example, consider {\em dog} $\models$ {\em animal} versus
{\em table} $\not\models$ {\em animal}. Suppose that {\em life} is one of the reference
words. We see that {\em dog} and {\em animal} are similar with respect to the reference word
{\em life}; the difference in their similarities is small. On the other hand,
{\em table} and {\em animal} are dissimilar with respect to {\em life}; there is
a large difference in their similarities. Some differences are important for
entailment (such as whether something is animate or inanimate) and others
usually have little effect (such as the colour of a thing). Given labeled training
data, we may be able to learn how differences in similarities affect lexical entailment.

We empirically evaluate the three algorithms, balAPinc, ConVecs, and SimDiffs, using three
different datasets. We find that SimDiffs performs significantly better than the other two
algorithms in some cases and there is no case for which it is significantly worse.
ConVecs is significantly worse than balAPinc and SimDiffs on one dataset, whereas
balAPinc is significantly worse than ConVecs on one dataset and significantly worse
than SimDiffs on two datasets.

Section~\ref{sec:defining} defines {\em lexical entailment} in terms of semantic relations 
between words. There is some disagreement about whether lexical entailment should be 
approached as a semantic relation classification task. We address this issue in
Section~\ref{sec:relations-entailment}. Past work on RLE is examined in Section~\ref{sec:related}.
Performance measures for RLE algorithms are presented in Section~\ref{sec:performance}. We 
describe the three algorithms in detail in Section~\ref{sec:algorithms}. The three algorithms 
are evaluated using three datasets, which are presented in Section~\ref{sec:datasets}. We use 
the datasets of \citeA{kotlerman10}, \citeA{baroni12}, and \citeA{jurgens12}. The experimental 
results are reported in Section~\ref{sec:experiments}. We discuss some implications of the 
experiments in Section~\ref{sec:discussion}. Limitations of this work are considered in
Section~\ref{sec:limits} and we conclude in Section~\ref{sec:conclusion}.

\section{Defining lexical entailment}
\label{sec:defining}

Let $w$ and $v$ be two words. \citeA[p. 442]{zhitomirsky09} define {\em substitutable 
lexical entailment} as follows:

\begin{quotation}

... $w$ entails $v$, if the following two conditions are fulfilled:

\begin{enumerate}

\item {\em Word meaning entailment}: the meaning of a possible sense of $w$ implies a
possible sense of $v$;

\item {\em Substitutability}: $w$ can substitute for $v$ in some naturally occurring sentence,
such that the meaning of the modified sentence would entail the meaning
of the original one.

\end{enumerate}

\end{quotation}

\noindent We call this the {\em substitutional definition} of lexical entailment.

We present a different definition of lexical entailment here. The idea is that whether one
word entails another depends on the semantic relation between the words. We discuss some
objections to this idea in Section~\ref{sec:relations-entailment}.

Let $x$ and $y$ be two words. To be able to say that $x$ entails $y$
outside of the context of a specific sentence, it must be the case that
there is a strong semantic relation between $x$ and $y$, and the entailment
must follow from the nature of that semantic relation. We say that $x$ entails $y$
if the following three conditions are fulfilled:

\begin{enumerate}

\item {\em Typical relation}: Given $x$ and $y$, there should be a typical semantic relation
$R(x,y)$ that comes to mind. Let $R(x,y)$ be the typical semantic relation between $x$
and $y$. If there is no typical semantic relation between $x$ and $y$, then $x$ cannot entail $y$
outside of a specific context.

\item {\em Semantic relation entailment}: If $x$ and $y$ typically have the semantic
relation $R(x,y)$, then it should follow from the meaning of the semantic relation
that $x$ implies $y$.

\item {\em Relation entailment agreement}: If $x$ and $y$ have two or more typical semantic
relations and the relations do not agree on whether $x$ implies $y$, then assume that $x$ does
not imply $y$.

\end{enumerate}

\noindent We call this the {\em relational definition} of lexical entailment.

In the first condition of the relational definition, the {\em typical relation} between $x$ 
and $y$ is the relation that naturally comes to mind when $x$ and $y$ are presented together. 
If $x$ and $y$ have multiple senses, the juxtaposition of $x$ and $y$ may suggest a semantic
relation and it may also constrain the possible senses of the words. The constrained senses
of the words are not necessarily the most frequent or prototypical senses of the words.

For example, consider the words {\em lion} and {\em cat}. The word {\em cat} has
the senses {\em house cat} (a specific type of cat) and {\em feline} (the general class of
cats, including domestic cats and wild cats). When the words {\em lion} and {\em cat}
are juxtaposed, the relation that naturally comes to mind (for us) is hyponym--hypernym 
(a {\em lion} is a type of {\em cat}) and the sense of {\em cat} is constrained to {\em feline}, 
although the {\em house cat} sense is more frequent and prototypical than the {\em feline} 
sense.

Context determines the sense of an ambiguous word, but lexical entailment considers
word pairs outside of the context of sentences. Since word senses can affect entailment,
any approach to lexical entailment must decide how to handle ambiguous words.
The substitutional definition of lexical entailment invites us to imagine a natural
sentence that provides the missing context and constrains the possible senses of the two words.
The relational definition of lexical entailment invites us to imagine a semantic
relation that connects the two words and constrains their possible senses.

The second condition of the relational definition determines whether one word entails 
another, based on their semantic relation. Since a hyponym implies its hypernym, {\em lion} 
entails {\em cat}. The second condition excludes semantic relations that do not imply 
entailment. For example, antonymy is excluded (e.g., {\em tall} does not imply {\em short})
and the hyponym--hypernym relation is only included when the direction is correct
(e.g., {\em lion} implies {\em cat} but {\em cat} does not imply {\em lion}).

The first condition in the substitutional definition of lexical entailment
(word meaning entailment) asks us to consider whether the sense of one word
implies the sense of another word. We hypothesize that any such implication must
depend on the semantic relation between the senses of the words. It seems to us that,
if there is no semantic relation between the words, then it is not possible for
one word to imply the other. If one words implies another, the implication must
follow from the nature of their semantic relation. The idea of the second condition
in the relational definition of lexical entailment is to make this connection between
semantic relations and lexical entailment explicit.

The third condition of the relational definition handles ambiguous cases by erring on the side 
of non-entailment. Some people might feel that {\em lion} and {\em cat} suggest either the 
hyponym--hypernym relation (assuming {\em cat} means {\em feline}) or the coordinate relation
(assuming that {\em cat} means {\em house cat}). Coordinates are words with a shared
hypernym. {\em Lion} and {\em house cat} share the hypernym {\em feline}. If
{\em cat} means {\em house cat}, then {\em lion} and {\em cat} are coordinates.
A hyponym implies its hypernym, but coordinates do not imply each other.
{\em Lion} implies {\em cat} in the {\em feline} sense but not in the {\em house cat}
sense. Thus these two relations (hyponym--hypernym and coordinate) do not agree on whether
{\em lion} implies {\em cat}. In this case, we believe that the hyponym--hypernym
is more natural, so we say that {\em lion} implies {\em cat}. For people who
feel both semantic relations are natural, the third condition says that there is no
entailment; for them, {\em lion} does not imply {\em cat}.

The third condition could be modified for different uses. For our dataset
(Section~\ref{subsec:jmth}), we chose to err on the side of non-entailment,
but ideally the choice would be made based on the downstream application. For some
applications, it may be better to err on the side of entailment. One possibility is
to give higher weight to some relations and use the weighting to choose between entailment
and non-entailment when two or more relations disagree. The weighting could be based
on the corpus frequency of the relations or the contexts in which the words appear.

To apply the relational definition of lexical entailment, it is helpful to have a taxonomy of
semantic relations, to provide options for $R$. In this paper, we use the taxonomy of \citeA{bejar91},
which includes seventy-nine subcategories of semantic relations, grouped into ten high-level
categories. The taxonomy is given in Tables~\ref{tab:bejar-relation-categories-1-5} and 
\ref{tab:bejar-relation-categories-6-10} in Section~\ref{subsec:jmth}.

It might seem that the relational definition redefines {\em lexical entailment} in a way
that makes our results incomparable with past results, but we believe that our results
are indeed comparable with past work. Both the substitutional definition and the relational 
definition are {\em operational definitions}: They are tests used to determine
the presence of entailment. They both require an understanding of the word {\em implies}, but
{\em implies} is a synonym for {\em entails}; they are not {\em theoretical definitions}
of entailment. They attempt to objectively capture the same underlying notion of
implication, and hence they may be compared and contrasted in terms of how well they
capture that notion.

\citeS{zhitomirsky09} substitutional definition of lexical entailment was intended
to capture only substitutional cases of entailment. They explicitly excluded non-substitutable
lexical entailment. They argue that their two conditions yield good inter-annotator agreement
and result in lexical entailment decisions that fit well with the needs of systems
for recognizing textual entailment.

We believe that there is a trade-off between inter-annotator agreement and coverage. The
substitutional and relational definitions differ regarding this trade-off. The
substitutional definition leads to higher levels of inter-annotator agreement than the 
relational definition, but the substitutional definition excludes (by design) important 
cases of lexical entailment (see Section~\ref{subsubsec:agreement}). 

Consider the following example:

\begin{verse}

Text: {\em Jane accidentally broke a glass.} \\*  
Hypothesis: {\em Jane accidentally broke something fragile.}

\end{verse}

\noindent For the word pair {\em glass:fragile}, the typical relation that comes
to mind is item:attribute, `an $x$ has the attribute $y$' (ID 5a in the semantic
relation taxonomy); thus the first condition of the relational definition is fulfilled.
An item entails its attributes; {\em glass} entails {\em fragile}; thus the second
condition is fulfilled. There are exceptions, such as bulletproof glass, but bulletproof
glass is not typical glass. There is no other typical relation between {\em glass} and
{\em fragile}, so the third condition is fulfilled.

One limitation of {\em substitutability} as defined by \citeA{zhitomirsky09} is
that it does not allow lexical entailment from one part of speech to another.
For example, {\em glass} entails {\em fragile}, but {\em glass} is a noun and
{\em fragile} is an adjective, so we cannot substitute one for the other in a
sentence. However, in spite of the difference in their parts of speech, it
seems reasonable to say that {\em glass} entails {\em fragile}. In a typical situation
that involves {\em glass}, the situation also involves something {\em fragile}.

As another example of a case where the substitutional definition excludes a
lexical entailment that the relational definition captures, consider {\em bequeath:heir},
an instance of the act:recipient relation (ID 7e in the relation taxonomy):

\begin{verse}

Text: {\em George bequeathed his estate to Jane.} \\*  
Hypothesis: {\em Jane was the heir to George's estate.}

\end{verse}

\noindent It is reasonable to say that the act of bequeathing entails that there is an heir,
although the verb {\em bequeathed} cannot be substituted for the noun {\em heir}.

To address this limitation of the substitutional definition, one possibility would be
to relax the definition of {\em substitutability} to cope with different parts of speech.
For example, given a noun $x$ and an adjective $y$, we could allow `an $x$' ({\em a glass})
to be replaced with `something $y$' ({\em something fragile}). Perhaps a relatively small
list of {\em substitutional patterns} could handle most part of speech substitution cases. However,
we do not pursue this option here, because it does not address a fundamental limitation of the
substitutional definition, which is the absence of semantic relations. We believe that
semantic relations and lexical entailment are intimately connected (see 
Section~\ref{sec:relations-entailment}).

The idea of substitional patterns suggests the generalization of lexical entailment
to phrasal entailment. For example, the phrase `$x$ bequeathed $y$ to $z$' entails the phrase
`$z$ was the heir to $x$'s $y$'. Patterns like this have been learned from corpora
\cite{lin01} and applied successfully to RTE \cite{mirkin09b}. However, our focus here
is lexical entailment, not phrasal entailment. We believe that a good algorithm for
lexical entailment should be useful as a component in an algorithm for phrasal entailment.

In our experiments, we use three different datasets. All three consist of word
pairs that have been labeled {\em entails} or {\em does not entail}. One dataset
(Section~\ref{subsec:kdsz}) was labeled using \citeS{zhitomirsky09} substitutional definition.
On preliminary inspection, it seems that the semantic relations in this dataset
are often part--whole and hyponym--hypernym relations, but the word pairs have not been systematically
labeled with relation categories. In another dataset (Section~\ref{subsec:bbds}), all of the pairs
that are labeled {\em entails} are instances of the hyponym--hypernym relation.
In the third dataset (Section~\ref{subsec:jmth}), the pairs were generated from
\citeS{bejar91} taxonomy. This dataset includes pairs sampled from all seventy-nine of the
subcategories in the taxonomy. Each pair was labeled {\em entails} or {\em does not
entail} based on the subcategory it came from. Tables~\ref{tab:bejar-relation-categories-1-5}
and \ref{tab:bejar-relation-categories-6-10} in Section~\ref{subsec:jmth} list
all of the subcategories of relations and their entailment labels.

Lexical entailment is sometimes asymmetric (e.g., for word pairs that are instances of the 
hyponym--hypernym relation) and sometimes symmetric (e.g., for synonyms) \cite{geffet05,kotlerman10}. 
Both the substitutional and relational definitions allow this blend
of symmetry and asymmetry.

In the semantic relation classification literature (discussed in Section~\ref{sec:related}),
supervised learning algorithms are applied to the task of classifying word pairs.
In general, these algorithms are capable of classifying both symmetric and asymmetric
relations. In particular, ConVecs and SimDiffs both approach lexical entailment
as a problem of supervised relation classification, and both are capable of learning
symmetric and asymmetric relations. They should be able to learn when lexical entailment
behaves asymmetrically (e.g., with cases like {\em glass:fragile}) and when it
behaves symmetrically (e.g., with cases like {\em car:automobile}).

The balAPinc measure is designed to capture asymmetry, but it is likely to give
approximately equal scores to {\em car:automobile} and {\em automobile:car}.
This can be seen by considering the details of its definition (see Section~\ref{subsec:balapinc}).

\section{Semantic relations and lexical entailment}
\label{sec:relations-entailment}

Some researchers have applied semantic relation classification to lexical entailment
\cite{akhmatova09,baroni12}, but \citeA[p. 443]{zhitomirsky09} have argued against this:

\begin{quotation}

... lexical entailment is not just a superset of other known relations,
but it is rather designed to select those sub-cases of other lexical relations that are needed
for applied entailment inference. For example, lexical entailment does not cover all cases
of meronyms (e.g., division does not entail company), but only some sub-cases of part--whole
relationship mentioned herein. In addition, some other relations are also covered
by lexical entailment, like ocean and water and murder and death, which do not seem to
directly correspond to meronymy or hyponymy relations.

Notice also that whereas lexical entailment is a directional relation that specifies
which word of the pair entails the other, the relation may hold in both directions
for a pair of words, as is the case for synonyms.

\end{quotation}

\noindent We agree with \citeA{zhitomirsky09} that some sub-cases of part--whole
involve lexical entailment and other sub-cases do not. However, this issue can
be addressed by breaking the part--whole category into subcategories.

One of the high-level categories in \citeS{bejar91} taxonomy is part--whole (ID 2 in the taxonomy), 
which has ten subcategories. We claim that eight of the ten subcategories involve entailment and 
two do not involve entailment, which is consistent with the claim that `lexical entailment does 
not cover all cases of meronyms' (in the above quotation).

Regarding `ocean and water and murder and death' (in the above quotation), the word pair
{\em ocean:water} is an instance of \citeS{bejar91} object:stuff subcategory (ID 2g in the
taxonomy) and {\em murder:death} is an instance of the cause:effect subcategory (ID 8a).
Regarding relations for which there is lexical entailment in both directions, synonymy (ID 3a) is
readily handled by marking it as entailing in both directions (see
Tables~\ref{tab:bejar-relation-categories-1-5} and \ref{tab:bejar-relation-categories-6-10}
in Section~\ref{subsec:jmth}).

We believe that \citeS{zhitomirsky09} argument is correct for high-level categories
but incorrect for subcategories. We offer the following hypothesis (introduced here):

\begin{verse}

{\em Semantic relation subcategories hypothesis:} Lexical entailment is {\em not} a superset of
high-level categories of semantic relations, but it {\em is} a superset of lower-level
subcategories of semantic relations.

\end{verse}

\noindent This hypothesis implies a tight connection between research in RLE and research
in semantic relation classification.

ConVecs and SimDiffs treat RLE as a semantic relation classification problem. These algorithms
do not require the semantic relation subcategories hypothesis: It is possible that it may be
fruitful to use ideas from research in semantic relation classification even if the hypothesis
is wrong. However, if the semantic relation subcategories hypothesis is correct, then there is
even more reason to treat RLE as a semantic relation classification problem.

We use the semantic relation subcategories hypothesis in Section~\ref{subsec:jmth},
as a new way of generating a dataset for evaluating RLE algorithms.
In our experiments (Section~\ref{sec:experiments}), we train the algorithms using data
based on \citeS{bejar91} taxonomy and then test them on previous lexical entailment datasets.

We do not claim that \citeS{bejar91} taxonomy handles all cases of lexical entailment, but
our results suggest that it covers enough cases to be effective. Future work may discover
lexical entailments that do not fit readily in \citeS{bejar91} taxonomy, but
we believe that the taxonomy can be expanded to handle exceptions as they are discovered.

\section{Related work}
\label{sec:related}

The first RTE Challenge took place in 2005 \cite{dagan06} and it has been a regular
event since then.\footnote{The RTE Challenge usually takes place once a year. See
the Textual Entailment Portal at \url{http://aclweb.org/aclwiki} for more information.}
Since the beginning, many RTE systems have included a module for recognizing lexical
entailment \cite{hickl06,herrera06}. The early RLE modules typically used a symmetric similarity
measure, such as the cosine measure \cite{salton83}, the LIN measure \cite{lin98b}, or
a measure based on WordNet \cite{pedersen04}, but it was understood that entailment is inherently
asymmetric and any symmetric measure can only be a rough approximation \cite{geffet05}.

\citeA{lillianlee99} proposed an asymmetric similarity measure for the degree to which a word
$a$ can be replaced by a word $b$ in a sentence, without substantially changing the meaning of
the sentence. \citeA{weeds03} introduced an asymmetric similarity measure for the degree to
which a specific term $a$ is subsumed by a more general term $b$ \cite<see also>{weeds04}.
This idea was developed further, specifically for application to lexical entailment, in a series
of papers that culminated in the balAPinc measure of the degree to which $a$ entails $b$
\cite{geffet05,szpektor08,zhitomirsky09,kotlerman10}. We describe balAPinc in detail in
Section~\ref{subsec:balapinc}.

\citeA{glickman06} define {\em lexical reference}, which is somewhat similar
to lexical entailment, but it is defined relative to a specific text, such as
a sentence. \citeA{mirkin09a} define {\em entailment between lexical elements},
which includes entailment between words and non-compositional elements. Their
definition is not based on substitutability; they accept many kinds of lexical
entailment that are excluded by substitutability. Their definition involves
what can be inferred from a lexical element in the context of some natural text. 

Compared to the number of papers on lexical entailment, there is a relatively large body
of literature on semantic relation classification
\cite{rosario01,rosario02,nastase03,turney06b,girju07}. Semantic relation classification
has been part of several SemEval (Semantic Evaluation) exercises:\footnote{See the
SemEval Portal at \url{http://aclweb.org/aclwiki} for more information.}

\begin{itemize}
  
\item SemEval-2007 Task 4: Classification of Semantic Relations between Nominals
\cite{girju07} \\
-- seven semantic relation classes

\item SemEval-2010 Task 8: Multi-Way Classification of Semantic Relations Between Pairs of
Nominals \cite{hendrickx10} \\
-- nine semantic relation classes

\item SemEval-2012 Task 2: Measuring Degrees of Relational Similarity \cite{jurgens12} \\
-- seventy-nine semantic relation classes

\end{itemize}

Only a few papers apply semantic relation classification to lexical
entailment \cite{akhmatova09,do10,baroni12,do12}. All of these papers emphasize
the hyponym--hypernym semantic relation, which is important for lexical entailment,
but it is not the only relation that involves entailment.

\citeA{baroni12} compared their ConVecs algorithm with the balAPinc measure and
found no significant difference in their performance. They also
consider how quantifiers (e.g., {\em some, all}) affect entailment.

Most algorithms for semantic relation classification are supervised
\cite{rosario01,rosario02,nastase03,turney06b,girju07}, although some are not
\cite{hearst92}. One objection to supervised learning for lexical entailment is
that it can require a large quantity of labeled training data.

\citeA{baroni12} offer an elegant solution to the training data issue, based on the
observation that, in adjective--noun phrases, the adjective--noun pair generally
entails the head noun. For example, {\em big cat} entails {\em cat}. This observation
allows them to label a large quantity of training data with relatively little effort.
However, their technique does not seem to be applicable to many of the relevant
subcategories in \citeS{bejar91} taxonomy. Our solution is to use word pairs that were
labeled with \citeS{bejar91} classes using Amazon's Mechanical Turk \cite{jurgens12}.
(See Section~\ref{subsec:jmth}.) This dataset covers a much wider range of semantic
relations than \citeS{baroni12} dataset.

\section{Performance measures}
\label{sec:performance}

One difference between an asymmetric similarity measure (such as balAPinc) and
a classification model based on supervised machine learning (such as ConVecs or
SimDiffs) is that the former yields a real-valued score whereas the latter gives a
binary-valued classification (0 = {\em does not entail} and 1 = {\em entails}). However,
this difference is superficial. Many supervised learning algorithms (including the algorithms
we use here) are able to generate a real-valued probability score (the probability that
the given example belongs in class 1). Likewise, it is easy to generate
a binary-valued class from a real-valued score by setting a threshold on the
score.

In our experiments (Section~\ref{sec:experiments}), we evaluate all three algorithms
both as real-valued asymmetric similarity measures and binary-valued classifiers. We
use average precision (AP) as a performance measure for real-valued scores, following
\citeA{kotlerman10}. We use precision, recall, F-measure, and accuracy as performance
measures for binary-valued classification, following \citeA{baroni12}.
The balAPinc measure (balanced average precision for distributional
inclusion) is partly inspired by the average precision measure,
thus it is useful to discuss average precision now, before we discuss
balAPinc (in Section~\ref{subsec:balapinc}).

\subsection{Average precision}
\label{subsec:ap}

AP was originally designed as a performance measure for information retrieval
systems. Suppose we have issued a query to a search engine and it has returned
a ranked list of $N$ documents, sorted in descending order of their automatically
estimated degree of relevance for our query. Assume that human judges have manually
labeled all of the documents as either {\em relevant} or {\em irrelevant} for the
given query. Let ${\rm P}(r)$ be the fraction of the top $r$ highest ranked
documents that have the label {\em relevant}. That is, ${\rm P}(r)$ is the
{\em precision} of the ranked list if we cut the list off after the $r$-th document.
Let ${\rm rel}(r)$ be 1 if the $r$-th document is labeled {\em relevant}, 0 otherwise.
AP is defined as follows \cite{buckley00}:

\begin{equation}
\label{eqn:ap}
{\rm AP} = \frac{\sum^{N}_{r=1} [{\rm P}(r) \cdot {\rm rel}(r)]}
{\mbox{total number of relevant documents}}
\end{equation}

\vspace{.25cm}

\noindent AP ranges from 0 (very poor performance) to 1 (perfect performance).
\citeA{buckley00} demonstrate that AP is more stable and more
discriminating than several alternative performance measures for information
retrieval systems.

The definition of AP reflects a bias in information retrieval. For a typical
query and a typical document collection, most documents are irrelevant and
the emphasis is on finding the few relevant documents. In machine learning,
if we have two classes, 0 and 1, they are usually considered equally important.
\citeA{kotlerman10} emphasize the class 1 ({\em entails}), but we believe
class 0 ({\em does not entail}) is also important. For example, the scoring
of the RTE Challenge gives an equal reward for recognizing when a text sentence entails
a hypothesis sentence and when it does not. Therefore we report two variations
of AP, which we call ${\rm AP}_0$ (average precision with respect to class 0)
and ${\rm AP}_1$ (average precision with respect to class 1), which we define
in the next paragraph.

Suppose we have a dataset of word pairs manually labeled 0 and 1. Let $N$ be the
number of word pairs in the dataset. Let ${\rm M}(a,b) \in \Re$ be a
measure that assigns a real-valued score to each word pair, $a\!:\!b$.
Sort the pairs in descending order of their ${\rm M}(a,b)$ scores.
Let ${\rm P}_1(r)$ be the fraction of the top $r$ highest ranked
pairs that have the label 1. Let ${\rm P}_0(r)$ be the fraction of the
bottom $r$ lowest ranked pairs that have the label 0. Let ${\rm C}_1(r)$
be 1 if the $r$-th document from the top is labeled 1, 0 otherwise.
Let ${\rm C}_0(r)$ be 1 if the $r$-th document from the bottom is labeled 0,
0 otherwise. Let $N_0$ be the total number of pairs labeled 0 and let $N_1$
be the total number of pairs labeled 1. We define ${\rm AP}_0$ and ${\rm AP}_1$
as follows:

\begin{align}
{\rm AP}_0 & = \frac{\sum^{N}_{r=1} [{\rm P}_0(r) \cdot {\rm C}_0(r)]}{N_0} \\
{\rm AP}_1 & = \frac{\sum^{N}_{r=1} [{\rm P}_1(r) \cdot {\rm C}_1(r)]}{N_1}
\end{align}

In their experiments, \citeA{kotlerman10} report only ${\rm AP}_1$.
It is possible to increase a system's performance according to ${\rm AP}_1$ at
the cost of lower ${\rm AP}_0$ performance. The formula for ${\rm AP}_1$
is more sensitive to the labels in the top of the list. What happens
at the bottom of the list has little impact on ${\rm AP}_1$, because
${\rm P}_1(r)$ gives a low weight to labels at the bottom of the list.
On the other hand, the formula for ${\rm AP}_0$ is more sensitive to labels
at the bottom of the list. If we focus on ${\rm AP}_1$ and ignore ${\rm AP}_0$,
we will prefer algorithms that get the top of the list right, even
if they do poorly with the bottom of the list. Therefore it is important to report
both ${\rm AP}_0$ and ${\rm AP}_1$.

\subsection{Precision, recall, F-measure, and accuracy}
\label{subsec:precision}

Like AP, precision and recall were originally designed as performance measures for
information retrieval systems. The precision of a system is an estimate of the conditional
probability that a document is truly relevant to a query, if the system says it is relevant.
The recall of a system is an estimate of the conditional probability that the system will say
that a document is relevant to a query, if it truly is relevant.

There is a tradeoff between precision and recall; one may be optimized at the
cost of the other. The F-measure is the harmonic mean of precision and recall. It
is designed to reward a balance of precision and recall.

Accuracy is a natural and intuitive performance measure, but it is sensitive
to the relative sizes of the classes. It is easy to interpret accuracy when
we have two equal-sized classes, but it is difficult to interpret when one
class is much larger than the other. The F-measure is a better measure when
the classes are not balanced.

As with AP, there are two variations of precision, recall, and F-measure, depending on whether
we focus on class 0 or class 1. Let $\mathbf{C}$ be a $2 \times 2$ confusion matrix,
where $c_{ij}$ is the number of word pairs that are actually in class $i$ and the
algorithm has predicted that they are in class $j$ (here $i,j \in \left\{ 0, 1 \right\}$).
We define precision, recall, and F-measure as follows:

{\allowdisplaybreaks 
\begin{align}
{\rm Pre}_0 & = c_{00} / (c_{00} + c_{10}) \\
{\rm Pre}_1 & = c_{11} / (c_{11} + c_{01}) \\
{\rm Rec}_0 & = c_{00} / (c_{00} + c_{01}) \\
{\rm Rec}_1 & = c_{11} / (c_{11} + c_{10}) \\
{\rm F}_0   & = 2 \cdot {\rm Pre}_0 \cdot {\rm Rec}_0 / ({\rm Pre}_0 + {\rm Rec}_0) \\
{\rm F}_1   & = 2 \cdot {\rm Pre}_1 \cdot {\rm Rec}_1 / ({\rm Pre}_1 + {\rm Rec}_1)
\end{align}
} 

\noindent Following standard practice \cite{witten11}, we merge the two variations of
each measure by taking their weighted averages, where the weights are determined
by the class sizes:

{\allowdisplaybreaks 
\begin{align}
w_0       & = (c_{00} + c_{01}) / (c_{00} + c_{01} + c_{10} + c_{11}) \\
w_1       & = (c_{11} + c_{10}) / (c_{00} + c_{01} + c_{10} + c_{11}) \\
{\rm Pre} & = w_0 \cdot {\rm Pre}_0 + w_1 \cdot {\rm Pre}_1 \\
{\rm Rec} & = w_0 \cdot {\rm Rec}_0 + w_1 \cdot {\rm Rec}_1 \\
{\rm F}   & = w_0 \cdot {\rm F}_0 + w_1 \cdot {\rm F}_1
\end{align}
} 

\noindent Finally, we define accuracy as usual:

\begin{equation}
{\rm Acc} = 100 \cdot (c_{00} + c_{11}) / (c_{00} + c_{01} + c_{10} + c_{11})
\end{equation}

\noindent The factor of 100 converts the accuracy from a fraction to a percentage score.

\section{Three approaches to lexical entailment}
\label{sec:algorithms}

In this section, we discuss the three approaches to RLE and describe the algorithms
for each approach in detail. All three approaches are based on word--context
matrices. For an introduction to the concepts behind word--context matrices,
see the survey paper by \citeA{turney10}.

In preliminary experiments with our development datasets, Dev1 and Dev2, we tuned the three
approaches to optimize their performance. We describe how Dev1 and Dev2 were generated in
Section~\ref{subsubsec:jmth-setup}. For each algorithm, we selected the matrix or matrices that
were most accurate with the development data. For both balAPinc and ConVecs, we chose the
word--context matrix from \citeA{turney11}. For SimDiffs, we chose two word--context matrices from
\citeA{turney12}.\footnote{Copies of all three matrices used here are available from the first
author by request.}

ConVecs and SimDiffs use support vector machines (SVMs) for supervised learning.
We used the development datasets to select the best kernels for the SVMs. The
best kernel for ConVecs was a second-degree polynomial kernel and the best kernel
for SimDiffs was a radial basis function (RBF) kernel. 

\subsection{The context inclusion hypothesis: balAPinc}
\label{subsec:balapinc}

We include balAPinc in our experiments because \citeA{kotlerman10} experimentally
compared it with a wide range of asymmetric similarity measures and found that
balAPinc had the best performance. The balAPinc asymmetric similarity measure is a
balanced combination of the asymmetric APinc measure \cite{kotlerman10} with the
symmetric LIN measure \cite{lin98b}. Balance is achieved by using the geometric mean:

\begin{equation}
\label{eqn:balAPinc}
{\rm balAPinc}(u, v) = \sqrt{ {\rm APinc}(u, v) \cdot {\rm LIN}(u,v) }
\end{equation}

To define APinc and LIN, we must first introduce some terminology. \citeA{kotlerman10}
define balAPinc with terminology from set theory, whereas ConVecs and SimDiffs
are more naturally defined with terminology from linear algebra. We will use
the set theoretical terminology of \citeA{kotlerman10} and the linear
algebraic terminology of \citeA{turney10}, so that the reader can easily see
both perspectives. This leads to a small amount of redundancy, but we believe
it is helpful to connect the two points of view.\footnote{ConVecs and SimDiffs
are fundamentally linear algebraic in conception, whereas balAPinc is fundamentally
set theoretic. We cannot readily describe all three systems with only one kind of
notation.}

First, some linear algebraic notation: Suppose that we have a word--context matrix, 
in which each row vector corresponds to a word and each column vector corresponds to a
context. Let $\mathbf{F}$ be the matrix of raw co-occurrence frequencies. If $w$ is the word
corresponding to the $i$-th row vector, $\mathbf{f}_{i:}$, and $c$ is the context
corresponding to the $j$-th column vector, $\mathbf{f}_{:j}$, then $f_{ij}$ is the number
of times $w$ occurs in the context $c$ in the given corpus.

Let the matrix $\mathbf{X}$ be the result of calculating the positive pointwise mutual
information (PPMI) between the word $w$ and the context $c$ for each element $f_{ij}$
in $\mathbf{F}$ \cite{bullinaria07,turney10}. PPMI takes the raw co-occurrence frequencies
and transforms them to weights that represent the importance of a given context for a given
word. The PPMI matrix $\mathbf{X}$ is typically sparse (most cells are zero) and no cells are
negative.\footnote{Other measures of word association may be used instead of PPMI.
See Chapter~5 of \citeA{manning99} for a good survey of association measures.}

The matrix $\mathbf{X}$ has the same number of rows ($n_r$) and columns ($n_c$) as the
raw frequency matrix $\mathbf{F}$. The value of an element $x_{ij}$ in $\mathbf{X}$ is
defined as follows \cite{turney10}:

{\allowdisplaybreaks 
\begin{align}
p_{ij} & = \frac{f_{ij}}{\sum_{i=1}^{n_r} \sum_{j=1}^{n_c} f_{ij}} \\
p_{i*} & = \frac{\sum_{j=1}^{n_c} f_{ij}}{\sum_{i=1}^{n_r} \sum_{j=1}^{n_c} f_{ij}} \\
p_{*j} & = \frac{\sum_{i=1}^{n_r} f_{ij}}{\sum_{i=1}^{n_r} \sum_{j=1}^{n_c} f_{ij}} \\
{\rm pmi}_{ij} & = \log \left ( \frac{p_{ij}}{p_{i*} p_{*j}} \right ) \\
x_{ij} & =
\left\{
\begin{array}{rl}
{\rm pmi}_{ij} & \mbox{if ${\rm pmi}_{ij} > 0$} \\
0 & \mbox{otherwise}
\end{array}
\right.
\end{align}
} 

Now, some set theoretical notation: Given a word $w$ corresponding to the $i$-th row
in $\mathbf{X}$, let $F_w$ be the set of contexts for which $x_{ij}$ is nonzero. That is,
$c \in F_w$ if and only if $x_{ij} \ne 0$, where $w$ corresponds to row $i$ and $c$
corresponds to column $j$. We may think of the contexts in the set $F_w$ as {\em features}
that characterize the word $w$. Let $|F_w|$ be the number of features in $F_w$. If $w$
corresponds to the $i$-th row in $\mathbf{X}$, then $|F_w|$ is the number of nonzero
cells in the $i$-th row vector, $\mathbf{x}_{i:}$.

Each feature $f$ in $F_w$ corresponds to a PPMI value $x_{ij}$. Let us rank the
features in $F_w$ in descending order of their corresponding PPMI values. Let
$f_{wr}$ be the $r$-th feature in the ranking of $F_w$, where $r$ ranges from 1 to $|F_w|$.
Let ${\rm rank}(f,F_w)$ be the rank of $f$ in $F_w$. Thus ${\rm rank}(f_{wr},F_w) = r$.
We want to normalize this rank so that it ranges between 0 and 1, where higher PPMI
values are closer to 1 and lower PPMI values are closer to 0. The function
${\rm rel}(f,F_w)$ provides this normalization:

\begin{equation}
\label{eqn:rel}
{\rm rel}(f,F_w) =
\left\{
\begin{array}{rl}
1 - \frac{{\rm rank}(f,F_w)}{|F_w| + 1} & \mbox{if $f \in F_w$} \\
0 & \mbox{if $f \notin F_w$}
\end{array}
\right.
\end{equation}

\noindent We may interpret ${\rm rel}(f,F_w)$ as a measure of the importance of the feature $f$
for characterizing the word $w$. This function is called {\em rel} because it is
somewhat analogous to {\em relevance} in information retrieval.

Recall the {\em context inclusion hypothesis}: If a word $u$ tends to occur in a subset of the
contexts in which a word $v$ occurs ($v$ contextually includes $u$), then $u$ (the
narrower term) tends to entail $v$ (the broader term). Suppose we test the features of $u$,
$f \in F_u$, in order of their rank, $r$, to see which features of $u$ are {\em contextually
included} in $v$. Let ${\rm inc}(r,F_u,F_v)$ be the set consisting of those features,
among the first $r$ features in $F_u$, that are included in $F_v$:

\begin{equation}
\label{eqn:inc}
{\rm inc}(r,F_u,F_v) = \left\{ f \, \middle| \, {\rm rank}(f,F_u) \le r \mbox{ and }
f \in (F_u \cap F_v) \right\}
\end{equation}

\noindent The size of this set, $|{\rm inc}(r,F_u,F_v)|$, ranges from 0 to $r$, where
$r \le |F_u|$. The function ${\rm P}(r,F_u,F_v)$ normalizes the size to range from 0 to 1:

\begin{equation}
\label{eqn:p}
{\rm P}(r,F_u,F_v) = \frac{|{\rm inc}(r,F_u,F_v)|}{r}
\end{equation}

\noindent We may interpret ${\rm P}(r,F_u,F_v)$ as a measure of the density
of $F_v$ features among the top $r$ features of $F_u$. This function is called {\em P}
because it is somewhat analogous to {\em precision} in information retrieval.

Now we are ready to define APinc:

\begin{equation}
\label{eqn:apinc}
{\rm APinc}(u, v) = \frac{\sum^{|F_u|}_{r=1} [{\rm P}(r,F_u,F_v) \cdot {\rm rel}(f_{ur},F_v)]}{|F_u|}
\end{equation}

\noindent APinc is a variation of the average precision (AP) measure, originally developed
for measuring the performance of information retrieval systems (see Section~\ref{subsec:ap}).
Consider the first term in the sum, $r = 1$. If $f_{u1}$, the highest-ranking feature in
$F_u$, is included in $F_v$, then ${\rm P}(1,F_u,F_v)$ will be 1; otherwise it will be 0. If
$f_{u1}$ is in $F_v$, then the product ${\rm P}(1,F_u,F_v) \cdot {\rm rel}(f_{u1},F_v)$
reduces to ${\rm rel}(f_{u1},F_v)$, the importance of the feature $f_{u1}$ for the word $v$.
APinc will have a high score when the most important features of $u$ are also
important features of $v$. APinc is asymmetric because it does not require that the
most important features of $v$ are important features of $u$.

Let $w_u(f)$ be the weight of the feature $f$ in the word $u$. The weight is given
by the PPMI value in $\mathbf{X}$. If $u$ corresponds to the $i$-th row and $f$
corresponds to the $j$-th column, then $w_u(f) = x_{ij}$. (It may seem redundant to
have both $w_u(f)$ and $x_{ij}$. The first is set theoretical and the second is
linear algebraic.) LIN is defined as follows \cite{lin98b}:

\begin{equation}
\label{eqn:lin}
{\rm LIN}(u,v) = \frac{\sum_{f \in F_u \cap Fv} [w_u(f) + w_v(f)]}
{\sum_{f \in F_u} w_u(f) + \sum_{f \in F_v} w_v(f)}
\end{equation}

\noindent In balAPinc (Equation~\ref{eqn:balAPinc}), the LIN measure is combined
with the APinc measure because the APinc measure by itself tends to be
sensitive to cases where $|F_u|$ or $|F_v|$ are unusually small \cite{kotlerman10}.

There are two parameters, ${\rm max}_F$ and $T$, that can be varied to control the
performance of balAPinc. The parameter ${\rm max}_F$ sets the maximum number of features
for each word. For a given word $w$, we calculate all of the features, $F_w$. If
$|F_w| > {\rm max}_F$, then we remove the lowest-ranking features until $|F_w| = {\rm max}_F$.
This reduces the impact of low-ranking features on the APinc score.
The parameter $T$ is a threshold for classification. If ${\rm balAPinc}(u, v) < T$,
then the word pair $u\!:\!v$ is classified as 0 ({\em does not entail}); otherwise,
$u\!:\!v$ is classified as 1 ({\em entails}). We describe how these parameters
are tuned in Section~\ref{sec:experiments}.

\citeA{kotlerman10} do not use the threshold $T$, since they do not evaluate
balAPinc as a classifier. They also do not use the parameter ${\rm max}_F$,
although their analysis supports the utility of this parameter; see Section~5.4.4
of \citeA{kotlerman10}.

In the experiments with balAPinc in Section~\ref{sec:experiments}, the PPMI matrix $\mathbf{X}$
is the same matrix as used by \citeA{turney11}. The matrix has 114,501 rows and 139,246 columns.
The rows correspond to single and multi-word entries ($n$-grams) in WordNet and the columns
correspond to unigrams in WordNet, distinguished according to whether they appear in the
left or right context of the given $n$-gram. The window size for context is four words
to the left and four words to the right of the $n$-gram. The matrix has a density (percentage
of nonzero values) of 1.22\%. 

The PPMI matrix is based on a corpus of $5 \times 10^{10}$ words, collected from
university websites by a webcrawler.\footnote{The corpus was collected by Charles
Clarke at the University of Waterloo.} The corpus was indexed with the Wumpus search
engine \cite{buettcher05}, which is designed for passage retrieval, rather
than document retrieval.\footnote{Wumpus is available at \url{http://www.wumpus-search.org/}.}
Suppose $f_{ij}$ is an element in the matrix of raw co-occurrence frequencies $\mathbf{F}$.
The $i$-th row of the matrix corresponds to an $n$-gram $w$ in WordNet and the $j$-th
column of the matrix corresponds to a unigram $c$. The value of $f_{ij}$ was calculated
by sending the query $w$ to Wumpus and counting the frequency of $c$ in the retrieved
passages. The matrix is described in detail in Section~2.1 of \citeA{turney11}.

It is common to smooth the PPMI matrix by applying a truncated singular value decomposition
(SVD) \cite{turney10}. On the development datasets, we experimented with smoothing
the matrix but the results were poor. The problem is that the truncated SVD yields a
matrix with a density of 100\%, but balAPinc is designed for highly sparse matrices.

Consider Equation~\ref{eqn:inc} for example. If the matrix has a density of 100\%,
then all of the contexts (all of the matrix columns) are nonzero, so $F_u$ and $F_v$
are simply the entire set of features, and $(F_u \cap F_v)$ is also the entire set
of features. Likewise, in Equation~\ref{eqn:lin}, all of the sums, $\sum_f$, range
over the entire set of features. The equations behind balAPinc are based on the
assumption that most of the elements in the matrix are zero (i.e., the matrix is sparse),
but this assumption is false if we apply a truncated SVD.

In the experiments in Section~\ref{sec:experiments}, we use the raw PPMI matrix,
with no SVD smoothing. \citeA{baroni12} also found that balAPinc works better without
SVD smoothing (see their Footnote 3).

\subsection{The context combination hypothesis: ConVecs}
\label{subsec:convecs}

With the ConVecs algorithm, \citeA{baroni12} were able to match the performance of
balAPinc. In ConVecs, we represent a word pair $a\!:\!b$ by the concatentation of the context
vectors $\mathbf{a}$ for $a$ and $\mathbf{b}$ for $b$. We apply a supervised learning
algorithm to a training set of word pairs, where each word pair is represented
by concatenated context vectors that are labeled {\em entails} or {\em does not entail}.
The supervised learning algorithm generates a classification model, which
enables us to assign labels to new word pairs, not present in the training data.

Let $\mathbf{X}$ be a word--context matrix, where the value of the cell $x_{ij}$ in $\mathbf{X}$
is given by the PPMI between the $i$-th word $w$ and the $j$-th context $c$. In our experiments,
we use the word--context matrix $\mathbf{X}$ from \citeA{turney11}, as in
Section~\ref{subsec:balapinc}, but now we smooth $\mathbf{X}$ with a truncated SVD.

SVD decomposes $\mathbf{X}$ into the product of three matrices
$\mathbf{U} \mathbf{\Sigma} \mathbf{V}^\mathsf{T}$, where $\mathbf{U}$ and $\mathbf{V}$ are
in column orthonormal form (i.e., the columns are orthogonal and have unit length,
$\mathbf{U}^\mathsf{T} \mathbf{U} = \mathbf{V}^\mathsf{T} \mathbf{V} = \mathbf{I}$)
and $\mathbf{\Sigma}$ is a diagonal matrix of singular values \cite{golub96}.
If $\mathbf{X}$ is of rank $r$, then $\mathbf{\Sigma}$ is also of rank $r$.
Let ${\mathbf{\Sigma}}_k$, where $k < r$, be the diagonal matrix formed from the top $k$
singular values, and let $\mathbf{U}_k$ and $\mathbf{V}_k$ be the matrices produced
by selecting the corresponding columns from $\mathbf{U}$ and $\mathbf{V}$. The matrix
$\mathbf{U}_k \mathbf{\Sigma}_k \mathbf{V}_k^\mathsf{T}$ is the matrix of rank $k$
that best approximates the original matrix $\mathbf{X}$, in that it minimizes the
approximation errors. That is, ${\bf \hat X} = \mathbf{U}_k \mathbf{\Sigma}_k \mathbf{V}_k^\mathsf{T}$
minimizes $\| {{\bf \hat X} - \mathbf{X}} \|_F$ over all matrices ${\bf \hat X}$ of rank $k$,
where $\| \ldots \|_F$ denotes the Frobenius norm \cite{golub96}.

We represent a word pair $a\!:\!b$ using row vectors from the matrix
$\mathbf{U}_k \mathbf{\Sigma}_k^p$. If $a$ and $b$ correspond to row vectors
$\mathbf{a}$ and $\mathbf{b}$ in $\mathbf{U}_k \mathbf{\Sigma}_k^p$, then $a\!:\!b$
is represented by the $2k$-dimensional vector that is the concatenation of $\mathbf{a}$
and $\mathbf{b}$. We normalize $\mathbf{a}$ and $\mathbf{b}$ to unit length before
we concatenate them.

There are two parameters in $\mathbf{U}_k \mathbf{\Sigma}_k^p$ that need to be set. The
parameter $k$ controls the number of latent factors and the parameter $p$ adjusts the
weights of the factors, by raising the corresponding singular values in $\mathbf{\Sigma}_k^p$
to the power $p$.

The parameter $k$ is well-known in the literature \cite{landauer07}, but $p$
is less familiar. \citeA{caron01} introduced $p$ for improving the performance of
truncated SVD with term--document matrices in information retrieval. The use of $p$ to
improve the performance with word--context matrices in lexical semantics is supported by
the empirical evaluations of \citeA{bullinaria12} and \citeA{turney12}. In the following
experiments (Section~\ref{sec:experiments}), we explore a range of values for $p$ and $k$.
\citeA{baroni12} use $k = 300$ and $p = 1$.\footnote{\citeA{baroni12} mention $k = 300$ in
their Footnote 3. In personal communication in November 2012, they said they used $p = 1$.}

Recall the {\em context combination hypothesis}: The tendency of $a$ to entail $b$ is
correlated with some learnable function of the contexts in which $a$ occurs and
the contexts in which $b$ occurs; some conjunctions of contexts tend to indicate
entailment and others tend to indicate a lack of entailment. Given the context combination 
hypothesis, vector concatenation is a natural way to represent $a\!:\!b$ for learning 
lexical entailment.

For their supervised learning algorithm, \citeA{baroni12} used Weka with LIBSVM.\footnote{Weka
is available at \url{http://www.cs.waikato.ac.nz/ml/weka/} and LIBSVM is available at
\url{http://www.csie.ntu.edu.tw/~cjlin/libsvm/}.} They used a polynomial kernel for
the support vector machine (SVM). We also use Weka and a polynomial kernel,
but we use the sequential minimal optimization (SMO) SVM in Weka \cite{platt98}, because it
can generate real-valued probability estimates, as well as binary-valued classes.
The probability estimates are based on fitting the outputs of the SVM with logistic
regression models \cite{witten11}.

We tried various kernels with ConVecs on the development datasets (Dev1 and Dev2; see
Section~\ref{subsubsec:jmth-setup}), and found that a second-degree polynomial kernel had
the best performance. We use the default settings for the polynomial kernel SMO SVM in Weka,
except we disable normalization, because the vectors are already normalized to the same length.

It seems to us that ConVecs is a good algorithm for a generic semantic relation,
but a representation that takes advantage of some background knowledge about
lexical entailment might require less training data. One thing we know about
lexical entailment is $a \models a$, for any $a$. ConVecs can only reliably
recognize that $a \models a$ if $a$ is similar to some $x$, such that the word pair
$x\!:\!x$ appears in the training data and has been labeled {\em entails}.
To cover a broad range of possible values for $a$, there must be many different
$x\!:\!x$ pairs in the training data. The ConVecs representation does not make
efficient use of the training data.

\subsection{The similarity differences hypothesis: SimDiffs}
\label{subsec:simdiffs}

SimDiffs uses two different word--context matrices, a {\em domain} matrix,
$\mathbf{D}$, and a {\em function} matrix, $\mathbf{F}$ \cite{turney12}. The domain
matrix is designed for measuring the domain similarity between two words (similarity
of topic, subject, or field). For example, {\em carpenter} and {\em wood} have a high
degree of domain similarity; they both come from the domain of {\em carpentry}.
The function matrix is designed for measuring function similarity (similarity of role,
relationship, or usage). For example, {\em carpenter} and {\em mason} have a high
degree of function similarity; they both function as {\em artisans}.

The two matrices use different types of context. The domain matrix uses the
{\em nouns} that occur near a given word as the context for the word, whereas the
function matrix uses the {\em verbs} that occur near the given word. The part-of-speech
information was generated with the OpenNLP tagger.\footnote{OpenNLP is available
at \url{http://opennlp.apache.org/}.} Our motivation for using two matrices in SimDiffs
is to generate a larger and more varied set of features for the supervised learning
algorithm. \citeA{turney12} demonstrated that domain and function matrices work together
synergetically when applied to semantic relations.

In experiments with the development datasets (Dev1 and Dev2), we tried using the domain
and function matrices with balAPinc and ConVecs, but both algorithms worked better with
the word--context matrix from \citeA{turney11}. For Sim\-Diffs, the combination of the
domain and function matrices from \citeA{turney12} had the best performance on the development
datasets.

Both $\mathbf{D}$ and $\mathbf{F}$ use PPMI and SVD, as in Section~\ref{subsec:convecs}.
This results in a total of four parameters that need to be tuned, $k_{\rm d}$ and
$p_{\rm d}$ for domain space and $k_{\rm f}$ and $p_{\rm f}$ for function space. In
the following experiments (Section~\ref{sec:experiments}), to simplify the search
through parameter space, we make $k_{\rm d} = k_{\rm f}$ and $p_{\rm d} = p_{\rm f}$.

The domain and function matrices are based on the same corpus as the word--context matrix
from \citeA{turney11}. Wumpus was used to index the corpus and search for passages, in the
same way as described in Section~\ref{subsec:balapinc}. $\mathbf{D}$ has 114,297 rows and
50,000 columns. The PPMI matrix has a density of 2.62\%. $\mathbf{F}$ has 114,101 rows and
50,000 columns. The PPMI matrix has a density of 1.21\%. For both matrices, truncated SVD
results in a density of 100\%.

The rows for both matrices correspond to single and multi-word entries ($n$-grams) in
WordNet. The columns are more complex; \citeA{turney12} provides a detailed description of
the columns and other aspects of the matrices. The matrices have different numbers of rows
because, before applying SVD, we removed rows that were entirely zero. The function matrix,
with its lower density, had more zero-valued rows than the domain matrix. 

Suppose that the words $a$ and $b$ correspond to the row vectors $\mathbf{a}_{\rm d}$ and
$\mathbf{b}_{\rm d}$ in the domain matrix, $\mathbf{D}$. Let ${\rm sim_d}(a,b)$ be the
similarity of $a$ and $b$ in domain space, as measured by the cosine of the angle
between $\mathbf{a}_{\rm d}$ and $\mathbf{b}_{\rm d}$,
${\rm cos}(\mathbf{a}_{\rm d},\mathbf{b}_{\rm d})$.
Likewise, let ${\rm sim_f}(a,b)$ be ${\rm cos}(\mathbf{a}_{\rm f},\mathbf{b}_{\rm f})$,
the cosine in function space.

Let $R$ be a set of reference words. Recall the {\em similarity differences hypothesis}:
The tendency of $a$ to entail $b$ is correlated with some learnable function of the 
differences in their similarities, ${\rm sim}(a,r) - {\rm sim}(b,r)$, to a set of reference 
words, $r \in R$; some differences tend to indicate entailment and others tend to indicate 
a lack of entailment. In Sim\-Diffs, we represent a word pair $a\!:\!b$ with a feature 
vector composed of four sets of features, $S_1$, $S_2$, $S_3$, and $S_4$, defined as follows:

{\allowdisplaybreaks 
\begin{align}
\label{eqn:s1} S_1 & = \left\{ {\rm sim_d}(a,r) - {\rm sim_d}(b,r) \, \middle| \, r \in R \right\} \\
\label{eqn:s2} S_2 & = \left\{ {\rm sim_f}(a,r) - {\rm sim_f}(b,r) \, \middle| \, r \in R \right\} \\
\label{eqn:s3} S_3 & = \left\{ {\rm sim_d}(a,r) - {\rm sim_f}(b,r) \, \middle| \, r \in R \right\} \\
\label{eqn:s4} S_4 & = \left\{ {\rm sim_f}(a,r) - {\rm sim_d}(b,r) \, \middle| \, r \in R \right\}
\end{align}
} 

\noindent $S_1$ is the difference between $a$ and $b$ in domain space, with
respect to their similarities to the reference words, $R$. $S_2$ is the difference
between $a$ and $b$ in function space. $S_1$ and $S_2$ are based on differences in
the same spaces, whereas $S_3$ and $S_4$ are based on differences in different spaces.

The cross-spatial differences ($S_3$ and $S_4$) may seem counterintuitive.
Consider the example {\em murder} $\models$ {\em death}, suggested by the
quotation from \citeA{zhitomirsky09} in Section~\ref{sec:relations-entailment}. Murder
typically involves two people, the victim and the aggressor, whereas death
typically involves one person, the deceased. This suggests that there
is a functional difference between the words, hence the function similarities
of {\em murder} may be quite different from the function similarities of {\em death}.
However, perhaps the domain similarities of {\em murder} are somewhat similar
to the function similarities of {\em death} ($S_3$) or perhaps the function
similarities of {\em murder} are somewhat similar to the domain similarities
of {\em death} ($S_4$). We include these similarities here to see if the
supervised learning algorithm can make use of them.

For $R$, the set of reference words, we use 2,086 words from Basic English
\cite{ogden30}.\footnote{This word list is available at
\url{http://ogden.basic-english.org/word2000.html}.}
Thus a word pair $a\!:\!b$ is represented by 2,086 $\times$ 4 = 8,344 features.
The words of Basic English were selected by \citeA{ogden30} to form a core vocabulary,
sufficient to represent most other English words by paraphrasing. We chose
this set of words because it is small enough to keep the number of features
manageable yet broad enough to cover a wide range of concepts. Other reference
words may also be suitable; this is a topic for future work.

We mentioned in Section~\ref{subsec:convecs} that ConVecs may be inefficient
for learning $a \models a$. On the other hand, consider how $a \models a$ is represented
in SimDiffs. Looking at Equations \ref{eqn:s1} and \ref{eqn:s2}, we see that, given
the word pair $a\!:\!a$, every feature in $S_1$ and $S_2$ will have the value zero.
Therefore it should not take many examples of $x\!:\!x$ in the training
data to learn that $a \models a$, for any $a$.

For our supervised learning algorithm, we use the SMO SVM in Weka.
Based on experiments with the development datasets (Dev1 and Dev2), we use
a radial basis function (RBF) kernel. We use the default settings, except we
disable normalization. We generate probability estimates for the classes.

\section{Three datasets for lexical entailment}
\label{sec:datasets}

This section describes the three datasets we use in our experiments. The first two
datasets have been used in the past for lexical entailment research. The third
dataset has been used for semantic relation research; this is the first time it has
been used for lexical entailment. We refer to each dataset by the initials
of the authors of the paper in which it was first reported.

\subsection{The KDSZ dataset}
\label{subsec:kdsz}

The KDSZ dataset was introduced by \citeA{kotlerman10} to evaluate balAPinc. The
dataset contains 3,772 word pairs, 1,068 labeled {\em entails} and 2,704 labeled
{\em does not entail}. It was created by taking a dataset of 3,200 labeled word pairs
from \citeA{zhitomirsky09} and adding 572 more labeled pairs.\footnote{Personal
communication with Zhitomirsky-Geffet in March 2012.} The labeling of the original
subset of 3,200 pairs is described in detail by \citeA{zhitomirsky09}. The definition
of lexical entailment that the judges used was the {\em substitutional definition}
given in Section~\ref{sec:defining}. Three judges labeled the pairs, with inter-annotator
agreement between any two of the three judges varying from 90.0\% to 93.5\%.

This dataset has two properties that complicate the experiments. First, the class
sizes are not balanced; 71.7\% of the pairs are labeled {\em does not entail}
and 28.3\% are labeled {\em entails}. Second, although every word pair is unique,
there are a few words that appear many times, in many different pairs. We address these
points in our experiments.

The words in the word pairs are mainly unigrams, but there are a few bigrams
({\em central bank, higher education, state government}). Fortunately all of the
bigrams appear in WordNet, so they have corresponding row vectors in our matrices.

\subsection{The BBDS dataset}
\label{subsec:bbds}

The BBDS dataset was created by \citeA{baroni12} and has been applied to evaluating
both balAPinc and ConVecs. In their paper, \citeA{baroni12} discuss several different
datasets. We use the dataset they call ${\rm N}_1 \models {\rm N}_2$, described in their
Section~3.3. The dataset contains 2,770 word pairs, 1,385 labeled {\em entails}
and 1,385 labeled {\em does not entail}. All of the 1,385 pairs labeled {\em entails}
are hyponym--hypernym noun--noun pairs, such as {\em pope} $\models$ {\em leader}.
The pairs were generated automatically from WordNet and then validated manually.

Although the class sizes are balanced, 50\% {\em entails} and 50\% {\em does not entail},
the BBDS dataset is not representative of the variety of semantic relations
that involve entailment, as we will see in Section~\ref{subsec:jmth}.
Also, although every word pair is unique, there are a few words that appear many times.
All of the word pairs are composed of unigrams and all of the unigrams appear
in WordNet, so they have corresponding row vectors in our matrices.

\subsection{The JMTH dataset}
\label{subsec:jmth}

\citeA{jurgens12} created a semantic relation dataset for SemEval-2012 Task~2:
Measuring Degrees of Relational Similarity.\footnote{The dataset is available at
\url{https://sites.google.com/site/semeval2012task2/}. We used the package called
SemEval-2012-Gold-Ratings.} This dataset contains 3,218 word pairs labeled with
seventy-nine types of semantic relations. In this section, we describe the original
SemEval-2012 dataset and the process we used to convert the dataset into 2,308 word
pairs, 1,154 labeled {\em entails} and 1,154 labeled {\em does not entail}.

The original dataset consists of word pairs labeled using the relation classification scheme of
\citeA{bejar91}. This is a hierarchical classification system with ten high-level categories,
each of which has between five and ten subcategories, for a total of seventy-nine
distinct subcategories.

For each subcategory in \citeS{bejar91} relation taxonomy, we have several types of
information, shown in Table~\ref{tab:info-types}. The first four types of information come
from \citeA{bejar91} and the rest were added by \citeA{jurgens12}.\footnote{All
of this information is provided in the file SemEval-2012-Complete-Data-Package at
\url{https://sites.google.com/site/semeval2012task2/download}.}

\begin{table}
\caption{Types of information about the classes in the relation taxonomy}
\begin{minipage}{\textwidth}
\begin{tabular}{lll}
\hline\hline
   & Type                                & Example \\
\hline
1. & ID                                  & 1a \\
2. & Category                            & class-inclusion \\
3. & Subcategory                         & taxonomic \\
4. & Paradigmatic examples ($x\!:\!y$)   & flower:tulip, emotion:rage, poem:sonnet \\
5. & Relational schema                   & $y$ is a kind/type/instance of $x$ \\
6. & Turker examples (Phase 1)           & fruit:grape, song:opera, rodent:mouse, ... \\
7. & Turker ratings (Phase 2)            & fruit:grape: 24.0, song:opera: -18.0, ... \\
\hline\hline
\end{tabular}
\end{minipage}
\label{tab:info-types}
\end{table}

The original SemEval-2012 dataset was generated in two phases, using Amazon's Mechanical
Turk \cite{jurgens12}.\footnote{See \url{https://www.mturk.com/}.} We refer to Mechanical
Turk workers as Turkers. In the first phase, for each of the seventy-nine
subcategories, Turkers were shown paradigmatic examples of word pairs in the
given subcategory, and they were asked to generate more word pairs of the same
semantic relation type. In the second phase, for each of the seventy-nine
subcategories, Turkers were shown word pairs that were generated in the first
phase, and they were asked to rate the pairs according to their degree
of prototypicality for the given semantic relation type. (See Table~\ref{tab:info-types}
for examples of the results of the two phases.)

We transformed the original SemEval-2012 semantic relation dataset to the new lexical
entailment dataset in four steps:

\begin{enumerate}

\item {\em Cleaning:} To improve the quality of the dataset, we removed the
ten lowest-rated word pairs from each subcategory. Since the original dataset
has 3,218 word pairs, the average subcategory has 40.7 word pairs. Our cleaning
operation reduced this to 30.7 pairs per subcategory, a total of 2,428 word pairs
($3218 - 79 \times 10 = 2428$).

\item {\em Doubling:} For each word pair $a\!:\!b$ labeled with a subcategory $X$,
we generated a new word pair $b\!:\!a$ and labeled it $X^{-1}$. For example,
{\em car:engine} is labeled object:component, so we created the pair {\em engine:car}
and labeled it ${\rm object\!:\!component}^{-1}$. This increased the number of
pairs to 4,856 and the number of subcategories to 158.

\item {\em Mapping:} We then mapped the 158 subcategory labels to the labels
0 ({\em does not entail}) and 1 ({\em entails}). The mapping is given in
Tables~\ref{tab:bejar-relation-categories-1-5} and \ref{tab:bejar-relation-categories-6-10}.
We assume all word pairs within a subcategory belong to the same class (either all
entail or none entail). (This assumption is tested in Section~\ref{subsubsec:agreement}.)
The result of mapping was 4,856 word pairs with two labels. There were 1,154 pairs
labeled 1 and 3,702 pairs labeled 0.

\item {\em Balancing:} To make a balanced dataset, we randomly removed pairs
labeled 0 until there were 1,154 pairs labeled 0 and 1,154 pairs labeled 1,
a total of 2,308 word pairs.

\end{enumerate}

Here is how to interpret Tables~\ref{tab:bejar-relation-categories-1-5} and
\ref{tab:bejar-relation-categories-6-10}: Given the pair {\em anesthetic:numbness} with
the label instrument:goal, we see from Table~\ref{tab:bejar-relation-categories-6-10}
(ID 8f) that $a \models b$ has the value 1, so we map the label instrument:goal
to the label 1 ({\em entails}). Given the pair {\em numbness:anesthetic} labeled
${\rm instrument\!:\!goal}^{-1}$, we see from the table (ID 8f) that $b \models a$
has the value 0, so we map the label ${\rm instrument\!:\!goal}^{-1}$ to the label 0
({\em does not entail}). In other words, {\em anesthetic} $\models$ {\em numbness}:

\begin{verse}

Text: {\em Due to the anesthetic, Jane felt no pain.}      \\*  
Hypothesis: {\em Due to the numbness, Jane felt no pain.}  \\*  
Text $\models$ Hypothesis

\end{verse}

\noindent However, {\em numbness} $\not\models$ {\em anesthetic}; for example, the numbness
might be caused by cold temperature, not anesthetic:

\begin{verse}

Text: {\em George had frostbite. Due to the numbness, he felt no pain.} \\*  
Hypothesis: {\em Due to the anesthetic, he felt no pain.}               \\*  
Text $\not\models$ Hypothesis

\end{verse}

Of the seventy-nine subcategories, twenty-five were labeled with class 1 for
$a \models b$ and twelve were labeled with class 1 for $b \models a$.
Note that two of the subcategories are labeled with class 1 for both
$a \models b$ and $b \models a$ (IDs 3a and 9e). This shows that our transformation
handles both symmetric and asymmetric semantic relations.

\begin{table}[H]   
\caption{Semantic relation categories 1 to 5, based on \citeA{bejar91}}
\begin{minipage}{\textwidth}
\begin{tabular}{llllcc}
\hline\hline
ID   & Category        & Subcategory           & Example ($a\!:\!b$)    & $a \models b$ & $b \models a$ \\
\hline
1a  & class-inclusion  & taxonomic             & flower:tulip           & 0  & 1 \\
1b  & class-inclusion  & functional            & weapon:knife           & 0  & 1 \\
1c  & class-inclusion  & singular:collective   & cutlery:spoon          & 0  & 1 \\
1d  & class-inclusion  & plural:collective     & dishes:saucers         & 0  & 1 \\
1e  & class-inclusion  & class:individual      & mountain:Everest       & 0  & 1 \\
2a  & part-whole       & object:component      & car:engine             & 1  & 0 \\
2b  & part-whole       & collection:member     & forest:tree            & 1  & 0 \\
2c  & part-whole       & mass:potion           & time:moment            & 1  & 0 \\
2d  & part-whole       & event:feature         & banquet:food           & 1  & 0 \\
2e  & part-whole       & stage:activity        & kickoff:football       & 0  & 1 \\
2f  & part-whole       & item:topological part & room:corner            & 1  & 0 \\
2g  & part-whole       & object:stuff          & glacier:ice            & 1  & 0 \\
2h  & part-whole       & creature:possession   & millionaire:money      & 1  & 0 \\
2i  & part-whole       & item:nonpart          & horse:wings            & 0  & 0 \\
2j  & part-whole       & item:ex-part          & prisoner:freedom       & 0  & 0 \\
3a  & similar          & synonymity            & car:auto               & 1  & 1 \\
3b  & similar          & dimension similarity  & simmer:boil            & 0  & 0 \\
3c  & similar          & dimension excessive   & concerned:obsessed     & 0  & 1 \\
3d  & similar          & dimension naughty     & listen:eavesdrop       & 0  & 1 \\
3e  & similar          & conversion            & grape:wine             & 0  & 0 \\
3f  & similar          & attribute similarity  & rake:fork              & 0  & 0 \\
3g  & similar          & coordinates           & son:daughter           & 0  & 0 \\
3h  & similar          & change                & crescendo:sound        & 1  & 0 \\
4a  & contrast         & contradictory         & alive:dead             & 0  & 0 \\
4b  & contrast         & contrary              & happy:sad              & 0  & 0 \\
4c  & contrast         & reverse               & buy:sell               & 0  & 0 \\
4d  & contrast         & directional           & left:right             & 0  & 0 \\
4e  & contrast         & incompatible          & slow:stationary        & 0  & 0 \\
4f  & contrast         & asymmetric contrary   & hot:cool               & 0  & 0 \\
4g  & contrast         & pseudoantonym         & popular:shy            & 0  & 0 \\
4h  & contrast         & defective             & limp:walk              & 1  & 0 \\
5a  & attribute        & item:attribute        & glass:fragile          & 1  & 0 \\
5b  & attribute        & attribute:condition   & edible:eaten           & 0  & 0 \\
5c  & attribute        & object:state          & beggar:poverty         & 1  & 0 \\
5d  & attribute        & attribute:state       & contentious:conflict   & 1  & 0 \\
5e  & attribute        & object:typical act    & soldier:fight          & 0  & 0 \\
5f  & attribute        & attribute:typical act & viable:live            & 0  & 0 \\
5g  & attribute        & act:act attribute     & creep:slow             & 1  & 0 \\
5h  & attribute        & act:object attribute  & sterilize:infectious   & 0  & 0 \\
5i  & attribute        & act:resultant         & rain:wet               & 1  & 0 \\
\hline\hline
\end{tabular}
\end{minipage}
\label{tab:bejar-relation-categories-1-5}
\end{table}

\begin{table}[H]   
\caption{Semantic relation categories 6 to 10, based on \citeA{bejar91}}
\begin{minipage}{\textwidth}
\begin{tabular}{llllcc}
\hline\hline
ID   & Category        & Subcategory           & Example ($a\!:\!b$)   & $a \models b$ & $b \models a$ \\
\hline
6a   & non-attribute   & item:nonattribute     & harmony:discordant    & 0  & 0 \\
6b   & non-attribute   & attr.:noncondition    & brittle:molded        & 0  & 0 \\
6c   & non-attribute   & object:nonstate       & laureate:dishonor     & 0  & 0 \\
6d   & non-attribute   & attr.:nonstate        & dull:cunning          & 0  & 0 \\
6e   & non-attribute   & obj.:atypical act     & recluse:socialize     & 0  & 0 \\
6f   & non-attribute   & attr.:atypical act    & reticent:talk         & 0  & 0 \\
6g   & non-attribute   & act:act nonattr.      & creep:fast            & 0  & 0 \\
6h   & non-attribute   & act:object nonattr.   & embellish:austere     & 0  & 0 \\
7a   & case relations  & agent:object          & tailor:suit           & 0  & 0 \\
7b   & case relations  & agent:recipient       & doctor:patient        & 0  & 0 \\
7c   & case relations  & agent:instrument      & farmer:tractor        & 0  & 0 \\
7d   & case relations  & act:object            & plow:earth            & 0  & 0 \\
7e   & case relations  & act:recipient         & bequeath:heir         & 1  & 0 \\
7f   & case relations  & object:recipient      & speech:audience       & 0  & 0 \\
7g   & case relations  & object:instrument     & pipe:wrench           & 0  & 0 \\
7h   & case relations  & recipient:instr.      & graduate:diploma      & 0  & 0 \\
8a   & cause-purpose   & cause:effect          & enigma:puzzlement     & 1  & 0 \\
8b   & cause-purpose   & cause:counteract      & hunger:eat            & 0  & 0 \\
8c   & cause-purpose   & enabler:object        & match:candle          & 0  & 0 \\
8d   & cause-purpose   & act:goal              & flee:escape           & 0  & 0 \\
8e   & cause-purpose   & agent:goal            & climber:peak          & 0  & 0 \\
8f   & cause-purpose   & instrument:goal       & anesthetic:numbness   & 1  & 0 \\
8g   & cause-purpose   & instrument:use        & abacus:calculate      & 0  & 0 \\
8h   & cause-purpose   & prevention            & pesticide:vermin      & 0  & 0 \\
9a   & space-time      & location:item         & arsenal:weapon        & 1  & 0 \\
9b   & space-time      & location:product      & bakery:bread          & 1  & 0 \\
9c   & space-time      & location:activity     & highway:driving       & 0  & 0 \\
9d   & space-time      & location:instr.       & beach:swimsuit        & 0  & 0 \\
9e   & space-time      & contiguity            & coast:ocean           & 1  & 1 \\
9f   & space-time      & time:activity         & childhood:play        & 0  & 0 \\
9g   & space-time      & time:associated       & retirement:pension    & 0  & 0 \\
9h   & space-time      & sequence              & prologue:narrative    & 0  & 0 \\
9i   & space-time      & attachment            & belt:waist            & 0  & 0 \\
10a  & reference       & sign:significant      & siren:danger          & 1  & 0 \\
10b  & reference       & expression            & hug:affection         & 1  & 0 \\
10c  & reference       & representation        & person:portrait       & 0  & 1 \\
10d  & reference       & plan                  & blueprint:building    & 0  & 1 \\
10e  & reference       & knowledge             & psychology:minds      & 1  & 0 \\
10f  & reference       & concealment           & disguise:identity     & 1  & 0 \\
\hline\hline
\end{tabular}
\end{minipage}
\label{tab:bejar-relation-categories-6-10}
\end{table}

\subsubsection{Making the mapping table}
\label{subsubsec:mapping}

We (Turney and Mohammad) each independently created a mapping like
Tables~\ref{tab:bejar-relation-categories-1-5} and \ref{tab:bejar-relation-categories-6-10}.
We disagreed on twelve of the 158 ($79 \times 2$) mappings (92.4\% agreement). We compared
our tables and discussed them until we arrived at a consensus. For all twelve disagreements,
our consensus was to label them 0. Tables~\ref{tab:bejar-relation-categories-1-5} and
\ref{tab:bejar-relation-categories-6-10} are the result of our consensus.

We used the first five types of information in Table~\ref{tab:info-types} to decide
how to map relation classes to entailment classes. Before we each independently
created a mapping table, we agreed to approach the task as follows:

\begin{quotation}

Procedure for annotation:

\begin{enumerate}

\item The relational schemas have more weight than the paradigmatic examples
when deciding whether $x$ entails $y$ or $y$ entails $x$.

\item Consider each of the paradigm pairs as instances of the given relational
schema. That is, interpret the pairs in the light of the schema. If the
three paradigmatic pairs are such that $x$ entails $y$, when interpreted this
way, then annotate the given category as `$x$ entails $y$', and likewise for
$y$ entails $x$. If two out of three paradigmatic pairs are such that $x$ entails
$y$, and the pair that is the exception seems unusual in some way, make a
note about the exceptional pair, for later discussion.

\item If any of the paradigmatic pairs are in the wrong order, correct their
order before proceeding. Make a note of the correction.

\end{enumerate}

\end{quotation}

\noindent We then compared our tables and combined them to form the final
Tables~\ref{tab:bejar-relation-categories-1-5} and \ref{tab:bejar-relation-categories-6-10}.

\subsubsection{Inter-annotator agreement}
\label{subsubsec:agreement}

As we mentioned above, we assume all word pairs within a subcategory belong to
the same class (either all entail or none entail). To test this assumption, we randomly
selected 100 word pairs, 50 labeled {\em entails} and 50 labeled {\em does not
entail}. We hid the labels and then we each independently manually labeled the pairs,
first using the relational definition of lexical entailment and then a second time
using the substitutional definition of lexical entailment (see Section~\ref{sec:defining}).
Table~\ref{tab:agreement} shows the percentage agreement between our manual labels and
automatic labeling, generated from the SemEval-2012 dataset by the mapping in
Tables~\ref{tab:bejar-relation-categories-1-5} and \ref{tab:bejar-relation-categories-6-10}.

\begin{table}
\caption{Agreement and entailment for the two definitions of lexical entailment}
\begin{minipage}{\textwidth}
\begin{tabular}{llcc}
\hline\hline
Measure            & Annotation               & Relational   & Substitutional  \\
\hline
percent agreement  & Turney versus Mohammad   &    81        &    89 \\
percent agreement  & Turney versus SemEval    &    81        &    68 \\
percent agreement  & Mohammad versus SemEval  &    70        &    65 \\
\noalign{\vspace{.25cm}}
percent entailing  & Turney                   &    51        &    22 \\
percent entailing  & Mohammad                 &    48        &    25 \\
percent entailing  & SemEval                  &    50        &    50 \\
\hline\hline
\end{tabular}
\end{minipage}
\label{tab:agreement}
\end{table}

With the relational definition of lexical entailment, we agreed on 81\% of the labels.
The agreement between our manual labels and the labels that were generated automatically,
by applying the mapping in Tables~\ref{tab:bejar-relation-categories-1-5} and
\ref{tab:bejar-relation-categories-6-10} to the SemEval dataset,
varied from 70\% to 81\%. These numbers suggest that our assumption that all word pairs
within a subcategory belong to the same class is reasonable. The assumption yields levels
of agreement that are comparable to the agreement in our manual labels.

We mentioned in Section~\ref{subsec:kdsz} that \citeA{zhitomirsky09} had inter-annotator
agreements in the 90\% range, whereas our agreement is 81\%. We hypothesize that
substitutability is a relatively objective test that leads to higher levels of agreement
but excludes important cases of lexical entailment. We discussed some examples of cases
that are missed by the substitutional definition in Section~\ref{sec:defining}.

Table~\ref{tab:agreement} shows that the agreement in our manual labels is 81\% for the
relational definition and 89\% for the substitutional definition. This supports our
hypothesis that substitutability is more objective. The agreement of 89\% is close to the
levels reported by \citeA{zhitomirsky09}. On the other hand, the number of pairs labeled
{\em entails} drops from 48-51\% for the relational definition to 22-25\% for the substitional
definition. This supports our hypothesis that substitutability excludes many cases of entailment.
The relational definition yields approximately twice the number of lexical entailments
that are captured by the substitutional definition.

As expected, the automated labeling using SemEval corresponds more closely
to manual labeling with the relational definition (70-81\%) than manual labeling
with the substitional definition (65-68\%). This confirms that the construction of
the dataset is in accordance with the intention of our relational definition.

\section{Experiments}
\label{sec:experiments}

In this section, we evaluate the three approaches to lexical entailment
(balAPinc, ConVecs, and SimDiffs) on the three datasets.

\subsection{Experiments with the JMTH dataset}
\label{subsec:jmth-exper}

For the first set of experiments, we used the JMTH dataset (Section~\ref{subsec:jmth}).
This dataset has 2,308 word pairs, 1,154 in class 0 and 1,154 in class 1.

\subsubsection{Experimental setup}
\label{subsubsec:jmth-setup}

For the experiments, we split the dataset into three (approximately) equal parts,
two development sets (Dev1 and Dev2) and one test set (Test).
The splits were random, except the balance of the class sizes was
maintained in all three subsets. Dev1 and Dev2 both contain 768 pairs
and Test contains 772 pairs.

Table~\ref{tab:jmth-categories-entailment} shows the number of word pairs in the Test
set for each of the ten high-level categories. In Tables~\ref{tab:bejar-relation-categories-1-5}
and \ref{tab:bejar-relation-categories-6-10},
we see that $a \models b$ is 0 for all subcategories of the category class-inclusion,
hence there are 0 pairs for $a \models b$ in the row for class-inclusion in
Table~\ref{tab:jmth-categories-entailment} but there are 12 pairs for $a \not\models b$.
On the other hand, in Tables~\ref{tab:bejar-relation-categories-1-5} and
\ref{tab:bejar-relation-categories-6-10}, $b \models a$ is 1
for all subcategories of the category class-inclusion, so it is not surprising
to see that there are 55 pairs for $b \models a$ in the row for class-inclusion in
Table~\ref{tab:jmth-categories-entailment} and 0 pairs for $b \not\models a$.
The number of pairs labeled {\em entails} is $261 + 125 = 386$ and the number
labeled {\em does not entail} is $176 + 210 = 386$.

\begin{table}
\caption{The distribution of the ten high-level categories in the Test set}
\begin{minipage}{\textwidth}
\begin{tabular}{rlrrrrr}
\hline\hline
ID & Category & $a \models b$ & $a \not\models b$ & $b \models a$ & $b \not\models a$ & Total \\
\hline
     1   & class-inclusion &   0   &   12   &   55  &     0   &   67 \\
     2   & part-whole      &  72   &    7   &   12  &    27   &  118 \\
     3   & similar         &  28   &   15   &   28  &    15   &   86 \\
     4   & contrast        &   9   &   25   &    0  &    26   &   60 \\
     5   & attribute       &  56   &   13   &    0  &    26   &   95 \\
     6   & non-attribute   &   0   &   40   &    0  &    31   &   71 \\
     7   & case relations  &   6   &   24   &    0  &    27   &   57 \\
     8   & cause-purpose   &  21   &   15   &    0  &    29   &   65 \\
     9   & space-time      &  35   &   17   &   10  &    17   &   79 \\
    10   & reference       &  34   &    8   &   20  &    12   &   74 \\
\noalign{\vspace{.25cm}}
    ---  & total           &  261  &  176   &  125  &   210   &  772 \\
\hline\hline
\end{tabular}
\end{minipage}
\label{tab:jmth-categories-entailment}
\end{table}

The balAPinc measure has two parameters to tune, ${\rm max}_F$ for the maximum
number of features and $T$ as a threshold for classification. On Dev1, we calculated
balAPinc five times, using five different values for ${\rm max}_F$, 1000, 2000, 3000,
4000, and 5000. For each given value of ${\rm max}_F$, we set $T$ to the value that
optimized the F-measure on Dev1. This gave us five pairs of values for ${\rm max}_F$ and $T$. We
tested each of these five settings on Dev2 and chose the setting that maximized the F-measure,
which was ${\rm max}_F = 1000$. The balAPinc measure is robust with respect to the parameter
settings. The accuracy on Dev2 ranged from 56.5\% with ${\rm max}_F = 1000$ to 52.5\% with
${\rm max}_F = 5000$. We kept the best ${\rm max}_F$ setting, but we tuned $T$ again on
the union of Dev1 and Dev2. With these parameter settings, we then applied balAPinc to
the Test set.

ConVecs has two parameters to tune, $k$ and $p$ for $\mathbf{U}_k \mathbf{\Sigma}_k^p$.
For $k$, we tried 100, 200, 300, 400, and 500. For $p$, we tried ten values, from 0.1
to 1.0 in increments of 0.1. For each of the fifty pairs of values for $k$ and $p$,
we ran Weka, using Dev1 as training data and Dev2 as testing data. The maximum F-measure
on Dev2 was achieved with $k = 100$ and $p = 0.4$. ConVecs is robust with respect
to the parameter settings. The accuracy on Dev2 ranged from a high of 70.1\% to
a low of 64.6\%. We then ran Weka one more time, using $k = 100$ and $p = 0.4$, with
the union of Dev1 and Dev2 as training data and Test as testing data.

SimDiffs has four parameters to tune, $k_{\rm d}$ and $p_{\rm d}$ for domain space
and $k_{\rm f}$ and $p_{\rm f}$ for function space, but we reduced this to two
parameters, $k$ and $p$, by setting $k = k_{\rm d} = k_{\rm f}$ and
$p = p_{\rm d} = p_{\rm f}$. We tried the same fifty pairs of values for $k$ and $p$
as we did with ConVecs. The maximum F-measure on Dev2 was achieved with $k = 200$ and
$p = 0.6$. SimDiffs is robust with respect to the parameter settings. The accuracy
on Dev2 ranged from 74.1\% to 68.2\%. We then ran Weka one more time, with the union of
Dev1 and Dev2 as training data and Test as testing data.

\subsubsection{Results}

Table~\ref{tab:jmth-three} shows the performance of all three algorithms on the Test
set. The accuracy of ConVecs (70.2\%) is not significantly different from the accuracy
of SimDiffs (72.4\%), according to Fisher's Exact Test \cite{agresti96}. However, both ConVecs and SimDiffs
are more accurate than balAPinc (57.3\%), at the 95\% confidence level. The other performance
measures (${\rm AP}_0$, ${\rm AP}_1$, Pre, Rec, and F) follow the same general pattern as
accuracy, which is what we would usually expect for a balanced dataset. The final column in
Table~\ref{tab:jmth-three} shows the 95\% confidence interval for accuracy, calculated
using the Wilson method.

\begin{table}
\caption{Comparison of the three algorithms on the JMTH dataset}
\begin{minipage}{\textwidth}
\begin{tabular}{lccccccc}
\hline\hline
Algorithm & ${\rm AP}_0$ & ${\rm AP}_1$ & Pre & Rec & F & Acc & 95\% C.I. \\
\hline
balAPinc   & 0.57 & 0.56 & 0.573 & 0.573 & 0.573 & 57.3 & 53.8--60.7 \\
ConVecs    & 0.76 & 0.77 & 0.703 & 0.702 & 0.702 & 70.2 & 66.9--73.3 \\
SimDiffs   & 0.80 & 0.79 & 0.724 & 0.724 & 0.724 & 72.4 & 69.1--75.4 \\
\hline\hline
\end{tabular}
\end{minipage}
\label{tab:jmth-three}
\end{table}

Table~\ref{tab:jmth-categories-accuracy} shows how the accuracies of the three algorithms
vary over the ten high-level categories in the Test set. ConVecs and SimDiffs have
roughly similar profiles but balAPinc is substantially different from the other two.
This is what we would expect, given that ConVecs and SimDiffs both approach
lexical entailment as a semantic relation classification problem, whereas balAPinc
approaches it as a problem of designing an asymmetric similarity measure. The
approach of balAPinc is near the level of the other two for some relation categories
(e.g., class-inclusion, non-attribute) but substantially below for others
(e.g., attribute, case relations, reference).

\begin{table}
\caption{Accuracy for each of the ten high-level categories in the Test set}
\begin{minipage}{\textwidth}
\begin{tabular}{rlcccr}
\hline\hline
ID & Category & balAPinc & ConVecs & SimDiffs & Size \\
\hline
     1   & class-inclusion & 79.1  & 76.1  &  88.1  &   67 \\
     2   & part-whole      & 61.0  & 69.5  &  66.9  &  118 \\
     3   & similar         & 46.5  & 44.2  &  54.7  &   86 \\
     4   & contrast        & 65.0  & 76.7  &  80.0  &   60 \\
     5   & attribute       & 44.2  & 68.4  &  70.5  &   95 \\
     6   & non-attribute   & 77.5  & 74.6  &  76.1  &   71 \\
     7   & case relations  & 43.9  & 64.9  &  71.9  &   57 \\
     8   & cause-purpose   & 49.2  & 73.8  &  75.4  &   65 \\
     9   & space-time      & 68.4  & 83.5  &  77.2  &   79 \\
    10   & reference       & 40.5  & 75.7  &  73.0  &   74 \\
\noalign{\vspace{.25cm}}
    ---  & average         & 57.3  & 70.2  &  72.4  &   77 \\
\hline\hline
\end{tabular}
\end{minipage}
\label{tab:jmth-categories-accuracy}
\end{table}

In Table~\ref{tab:jmth-subsets}, we explore the contribution of each set of features
to the performance of SimDiffs. In the columns for $S_1$ to $S_4$, a value of 1
indicates that the set is included in the feature vector and 0 indicates that the set
is excluded (see Section~\ref{subsec:simdiffs}). $S_1$ is the difference between
$a$ and $b$ in domain space, with respect to their similarities to the reference words,
$R$. $S_2$ is the difference between $a$ and $b$ in function space. $S_1$ and $S_2$ are
based on differences in the same spaces, whereas $S_3$ and $S_4$ are based on differences
in different spaces. The parameters are tuned individually for each row in Table~\ref{tab:jmth-subsets},
the same way they are tuned for SimDiffs in Table~\ref{tab:jmth-three}. The results are based
on the Test set.

\begin{table}
\caption{Experiments with subsets of SimDiffs features}
\begin{minipage}{\textwidth}
\begin{tabular}{ccccccccccc}
\hline\hline
$S_1$ & $S_2$ & $S_3$ & $S_4$ & ${\rm AP}_0$ & ${\rm AP}_1$ & Pre & Rec & F & Acc & 95\% C.I. \\
\hline
1     &  1    &  1    &  1    & 0.80 & 0.79 & 0.724 & 0.724 & 0.724 & 72.4 & 69.1--75.4 \\
\noalign{\vspace{.25cm}}
1     &  1    &  0    &  0    & 0.76 & 0.75 & 0.680 & 0.680 & 0.680 & 68.0 & 64.6--71.2 \\
0     &  0    &  1    &  1    & 0.79 & 0.79 & 0.717 & 0.716 & 0.716 & 71.6 & 68.3--74.7 \\
\noalign{\vspace{.25cm}}
1     &  0    &  0    &  0    & 0.71 & 0.69 & 0.663 & 0.663 & 0.663 & 66.3 & 62.9--69.6 \\
0     &  1    &  0    &  0    & 0.75 & 0.72 & 0.684 & 0.684 & 0.684 & 68.4 & 65.0--71.6 \\
0     &  0    &  1    &  0    & 0.76 & 0.74 & 0.690 & 0.690 & 0.690 & 69.0 & 65.7--72.2 \\
0     &  0    &  0    &  1    & 0.75 & 0.73 & 0.701 & 0.701 & 0.701 & 70.1 & 66.8--73.2 \\
\hline\hline
\end{tabular}
\end{minipage}
\label{tab:jmth-subsets}
\end{table}

Most of the differences in the accuracies in Table~\ref{tab:jmth-subsets} are
not significant, but the accuracy of all of the features together (72.4\%) is
significantly higher than the accuracy of $S_1$ and $S_2$ without the help
of $S_3$ and $S_4$ (68.0\%), according to Fisher's Exact Test at the 95\% confidence
level. This supports the view that working with two different spaces has a synergetic
effect, since each feature in $S_3$ and $S_4$ is based on two different spaces,
whereas each feature in $S_1$ and $S_2$ is based on one space. (See the discussion
of this in Section~\ref{subsec:simdiffs}.)

Let {\em Gen} (general) refer to the matrix from \citeA{turney11} and let {\em Dom}
and {\em Fun} refer to the domain and function matrices from \citeA{turney12}. In
Section~\ref{sec:algorithms}, we mentioned that we performed experiments on the
development datasets (Dev1 and Dev2) in order to select the matrices for each
algorithm. Based on these experiments, we chose the Gen matrix for both
balAPinc and ConVecs, and we chose the Dom and Fun matrices for SimDiffs.

In Table~\ref{tab:jmth-matrices}, we vary the matrices and evaluate the performance
on the Test set, to see whether the development datasets were a reliable guide for
choosing the matrices. The matrices that were chosen based on the development datasets are
in bold font. For balAPinc, Gen (57.3\%) is indeed the best matrix. For ConVecs,
it seems that Fun (71.9\%) might be a better choice than Gen (70.2\%), but the difference
in their accuracy is not statistically significant. For SimDiffs, Dom and Fun
(72.4\%) are slightly less accurate than Gen and Fun (72.8\%), but again the
difference is not significant. As expected, no matrices are significantly better on the
Test set than the matrices that were chosen based on the development datasets.

\begin{table}
\caption{Effect of the matrices on the performance of the algorithms}
\begin{minipage}{\textwidth}
\begin{tabular}{llccccccc}
\hline\hline
Algorithm & Matrices & ${\rm AP}_0$ & ${\rm AP}_1$ & Pre & Rec & F & Acc & 95\% C.I. \\
\hline
balAPinc   & {\bf Gen}      & 0.57 & 0.56 & 0.573 & 0.573 & 0.573 & 57.3 & 53.8--60.7 \\
           & Dom            & 0.53 & 0.54 & 0.532 & 0.532 & 0.532 & 53.2 & 49.7--56.7 \\
           & Fun            & 0.54 & 0.53 & 0.530 & 0.530 & 0.530 & 53.0 & 49.5--56.5 \\
\noalign{\vspace{.25cm}}
ConVecs    & {\bf Gen}      & 0.76 & 0.77 & 0.703 & 0.702 & 0.702 & 70.2 & 66.9--73.3 \\
           & Dom            & 0.72 & 0.75 & 0.676 & 0.675 & 0.674 & 67.5 & 64.1--70.7 \\
           & Fun            & 0.79 & 0.78 & 0.719 & 0.719 & 0.719 & 71.9 & 68.6--75.0 \\
\noalign{\vspace{.25cm}}
SimDiffs   & {\bf Dom, Fun} & 0.80 & 0.79 & 0.724 & 0.724 & 0.724 & 72.4 & 69.1--75.4 \\
           & Gen, Fun       & 0.79 & 0.79 & 0.728 & 0.728 & 0.728 & 72.8 & 69.5--75.9 \\
           & Dom, Gen       & 0.77 & 0.77 & 0.702 & 0.702 & 0.702 & 70.2 & 66.9--73.3 \\
           & Gen, Gen       & 0.75 & 0.76 & 0.689 & 0.689 & 0.689 & 68.9 & 65.6--72.1 \\
\hline\hline
\end{tabular}
\end{minipage}
\label{tab:jmth-matrices}
\end{table}

\subsection{Experiments with the KDSZ dataset}
\label{subsec:kdsz-exper}

The second set of experiments used the KDSZ dataset (Section~\ref{subsec:kdsz}).
This dataset has 3,772 word pairs, 2,704 in class 0 and 1,068 in class 1.

\subsubsection{Experimental setup}

We experimented with four different ways of splitting the dataset. The {\em Evaluation}
column in Table~\ref{tab:kdsz-three} indicates the experimental setup (dataset splitting).

The {\em standard} evaluation is ten-fold cross-validation
in which the folds are random. This evaluation yields relatively high scores,
because, although every pair in the KDSZ dataset is unique, many pairs share a
common term. This makes supervised learning easier, because a pair in the testing
fold will often share a term with several pairs in the training folds.

The {\em clustered} evaluation is designed to be more challenging than the {\em standard}
evaluation. The {\em clustered} evaluation is ten-fold cross-validation with
non-random folds. We put pairs that share a common term into the same fold. Due
to the large number of pairs with shared terms, it is not possible to construct
ten folds such that there are absolutely no terms that are shared by any two folds. Therefore
we gave a high priority to isolating the most common shared words to single folds, but
we allowed a few less common shared words to appear in more than one fold. Thus
a pair in the testing fold will only rarely share a term with pairs in the training folds.

The {\em standard} and {\em clustered} evaluations have more examples in
class 0 ({\em does not entail}) than in class 1 ({\em entails}). The {\em balanced}
dataset takes the {\em clustered} evaluation a step further, by first clustering folds
and then randomly removing pairs labeled as class 0, until the folds all have an equal
number of pairs in both classes.

For the {\em different} evaluation, instead of cross-validation, the algorithms are
trained on the JMTH dataset and tested on the KDSZ dataset, after the KDSZ dataset has
been balanced by randomly removing pairs labeled as class 0.

The balAPinc measure has two parameters, ${\rm max}_F$ for the maximum
number of features and $T$ as a threshold for classification. In all four
experimental setups, we used the setting ${\rm max}_F = 1000$, based on
the tuning experiments with the JMTH dataset (Section~\ref{subsec:jmth-exper}).
For $T$, we used the training split in each of the four experimental setups.
For the {\em standard}, {\em clustered}, and {\em balanced} setups, the training
split is the nine folds used for training in each step of the ten-fold cross-validation.
For the {\em different} setup, the training split is the whole JMTH dataset.
For all four setups, we set $T$ to the value that optimized the F-measure on the training split.

ConVecs has two parameters to tune, $k$ and $p$ for $\mathbf{U}_k \mathbf{\Sigma}_k^p$.
In all four experimental setups, we used $k = 100$ and $p = 0.4$,
based on the experiments with the JMTH dataset. The training splits
were used to teach the supervised learning algorithm (the polynomial
kernel SMO SVM in Weka).

SimDiffs has four parameters to tune. We used $k_{\rm d} = k_{\rm f} = 200$ and
$p_{\rm d} = p_{\rm f} = 0.6$, based on the experiments with the JMTH dataset.
The training splits were used to teach the supervised learning algorithm (the RBF
kernel SMO SVM in Weka).

\subsubsection{Results}

In Table~\ref{tab:kdsz-three}, the four experimental setups (standard, clustered,
balanced, and different) are given in order of increasing challenge and increasing
realism. Of the four experimental setups, we believe that the {\em different} evaluation
is the most challenging and most realistic. If an RLE module is part of a commercial RTE
system, the module will inevitably encounter word pairs in the field that are quite
different from the pairs it saw during training. The {\em different} evaluation comes
closest to approximating field usage.

\begin{table}
\caption{Comparison of the three algorithms on the KDSZ dataset}
\begin{minipage}{\textwidth}
\begin{tabular}{llccccccc}
\hline\hline
Algorithm & Evaluation & ${\rm AP}_0$ & ${\rm AP}_1$ & Pre & Rec & F & Acc & 95\% C.I. \\
\hline
balAPinc   & standard    & 0.79 & 0.37 & 0.645 & 0.645 & 0.645 & 64.5 & 63.0--66.0 \\
           & clustered   & 0.79 & 0.37 & 0.644 & 0.643 & 0.644 & 64.3 & 62.8--65.8 \\
           & balanced    & 0.60 & 0.59 & 0.583 & 0.583 & 0.583 & 58.3 & 56.2--60.4 \\
           & different   & 0.61 & 0.60 & 0.582 & 0.582 & 0.582 & 58.2 & 56.1--60.3 \\
\noalign{\vspace{.25cm}}
ConVecs    & standard    & 0.87 & 0.56 & 0.731 & 0.747 & 0.735 & 74.7 & 73.3--76.1 \\
           & clustered   & 0.78 & 0.36 & 0.636 & 0.690 & 0.645 & 69.0 & 67.5--70.5 \\
           & balanced    & 0.60 & 0.59 & 0.567 & 0.554 & 0.531 & 55.4 & 53.3--57.5 \\
           & different   & 0.57 & 0.62 & 0.569 & 0.561 & 0.547 & 56.1 & 54.0--58.2 \\
\noalign{\vspace{.25cm}}
SimDiffs   & standard    & 0.88 & 0.60 & 0.749 & 0.757 & 0.752 & 75.7 & 74.3--77.0 \\
           & clustered   & 0.80 & 0.40 & 0.664 & 0.684 & 0.671 & 68.4 & 66.9--69.9 \\
           & balanced    & 0.63 & 0.64 & 0.596 & 0.592 & 0.588 & 59.2 & 57.1--61.3 \\
           & different   & 0.58 & 0.61 & 0.581 & 0.574 & 0.564 & 57.4 & 55.3--59.5 \\
\hline\hline
\end{tabular}
\end{minipage}
\label{tab:kdsz-three}
\end{table}

On the {\em different} evaluations, balAPinc achieves an accuracy of 58.2\%, ConVecs
has an accuracy of 56.1\%, and SimDiffs reaches 57.4\%. There is no statistically
significant difference between any of these accuracies, according to Fisher's Exact
Test at the 95\% confidence level.

With ConVecs and SimDiffs, compared to balAPinc, there is a relatively large
gap between the {\em standard} performance and the {\em different} performance.
This is because ConVecs and SimDiffs use supervised learning and thus they
benefit from the {\em standard} setup, where the training data is highly similar
to the testing data. In balAPinc, the training data is used only to tune the
threshold, $T$, which limits the benefit of the training data.

Note that the gap between the {\em standard} performance and the {\em different}
performance is not simply a question of the quantity of data. In the
{\em different} setup, there is a qualitative difference between the training
data and the testing data. Increasing the size of the training dataset with
more data of the same type will not be helpful. The goal of the {\em different}
setup is to test the ability of the algorithms to bridge the qualitative
gap between the training and testing data. This qualitative gap is more
challenging for supervised learning than a quantitative gap. It is a gap that
learning algorithms inevitably face in real applications \cite{pan10}.

The KDSZ dataset has been used in previous research, but the past results are
not comparable with our results. \citeA{kotlerman10} reported ${\rm AP}_1$ without
${\rm AP}_0$, but there is a trade-off between ${\rm AP}_1$ and ${\rm AP}_0$.
\citeA{kotlerman10} did not attempt to evaluate balAPinc as a classifier,
so they did not report precision, recall, F-measure, or accuracy.

\subsection{Experiments with the BBDS dataset}
\label{subsec:bbds-exper}

The final set of experiments used the BBDS dataset (Section~\ref{subsec:bbds}).
The dataset has 2,770 word pairs, 1,385 in class 0 and 1,385 in class 1.

\subsubsection{Experimental setup}

We experimented with three different ways of splitting the dataset.
In Table~\ref{tab:bbds-three}, the evaluations follow the same setups as in
Table~\ref{tab:kdsz-three}. However, there is no {\em balanced} setup, since
the BBDS dataset is already balanced. In the {\em different} evaluation,
the algorithms are trained on the JMTH dataset and evaluated on the BBDS.
This is the most realistic evaluation setup.

\subsubsection{Results}

In Table~\ref{tab:bbds-three}, on the {\em different} evaluations, balAPinc achieves
an accuracy of 68.7\%, ConVecs has an accuracy of 65.1\%, and SimDiffs reaches 74.5\%.
All of these accuracies are significantly different, according to Fisher's Exact Test
at the 95\% confidence level.

\begin{table}
\caption{Comparison of the three algorithms on the BBDS dataset}
\begin{minipage}{\textwidth}
\begin{tabular}{llccccccc}
\hline\hline
Algorithm & Evaluation & ${\rm AP}_0$ & ${\rm AP}_1$ & Pre & Rec & F & Acc & 95\% C.I. \\
\hline
balAPinc   & standard    & 0.79 & 0.73 & 0.722 & 0.722 & 0.722 & 72.2 & 70.5--73.8 \\
           & clustered   & 0.79 & 0.73 & 0.722 & 0.722 & 0.722 & 72.2 & 70.5--73.8 \\
           & different   & 0.79 & 0.73 & 0.701 & 0.687 & 0.682 & 68.7 & 67.0--70.4 \\
\noalign{\vspace{.25cm}}
ConVecs    & standard    & 0.95 & 0.95 & 0.876 & 0.876 & 0.876 & 87.6 & 86.3--88.8 \\
           & clustered   & 0.92 & 0.91 & 0.829 & 0.821 & 0.819 & 82.1 & 80.6--83.5 \\
           & different   & 0.72 & 0.71 & 0.652 & 0.651 & 0.650 & 65.1 & 63.3--66.9 \\
\noalign{\vspace{.25cm}}
SimDiffs   & standard    & 0.97 & 0.97 & 0.913 & 0.913 & 0.913 & 91.3 & 90.2--92.3 \\
           & clustered   & 0.96 & 0.96 & 0.883 & 0.881 & 0.881 & 88.1 & 86.8--89.3 \\
           & different   & 0.84 & 0.82 & 0.751 & 0.745 & 0.743 & 74.5 & 72.8--76.1 \\
\hline\hline
\end{tabular}
\end{minipage}
\label{tab:bbds-three}
\end{table}

The BBDS data was used by \citeA{baroni12} to compare balAPinc
with ConVecs. They used two different evaluation setups, similar to our {\em standard}
and {\em different} setups. For balAPinc using a {\em standard} setup, they obtained an
accuracy of 70.1\%, slighly below our result of 72.2\%. The difference is likely due to
minor differences in the word--context matrices that we used. For balAPinc using a
{\em different} setup, their accuracy was 70.4\%, compared to our 68.7\%. They used their
own independent dataset to tune balAPinc, whereas we used the JMTH
dataset. Given that our word--context matrices and our training data are different
from theirs, the accuracies are closer than might be expected.\footnote{These
accuracy numbers and the numbers reported in the next paragraph are taken from
Table 2 in \citeA{baroni12}.}

For ConVecs using a {\em standard} setup, \citeA{baroni12}
report an accuracy of 88.6\%, whereas we achived 87.6\%. Using a {\em different} setup,
they obtained 69.3\%, whereas our accuracy was 65.1\%. It seems likely that
our training data (the JMTH dataset) was less similar to the BBDS dataset
than their own independent dataset, which made our {\em different} setup more
challenging than theirs. Nonetheless, the accuracies are closer than might be expected,
given the differences in the setups.

\section{Discussion of results}
\label{sec:discussion}

Table~\ref{tab:summary1} summarizes the accuracy results from the experiments.
For the KDSZ and BBDS experiments, only the {\em different} evaluation is shown.
Bold font is used to mark the cases where the accuracy is significantly less than
the accuracy of SimDiffs. In no case is the accuracy significantly greater than
the accuracy of SimDiffs.

\begin{table}
\caption{Summary of accuracy for the three algorithms with the three datasets}
\begin{minipage}{\textwidth}
\begin{tabular}{lccc}
\hline\hline
Algorithm  & JMTH Accuracy & KDSZ Accuracy & BBDS Accuracy \\
\hline
balAPinc   & {\bf 57.3}    & 58.2          & {\bf 68.7} \\
ConVecs    &      70.2     & 56.1          & {\bf 65.1} \\
SimDiffs   &      72.4     & 57.4          &      74.5  \\
\hline\hline
\end{tabular}
\end{minipage}
\label{tab:summary1}
\end{table}

The JMTH dataset is based on seventy-nine types of semantic relations.
The pairs in this dataset were labeled in accordance with the relational
definition of lexical entailment (see Section~\ref{sec:defining}). This explains
why balAPinc, which was designed with the substitutional definition in mind, performs
poorly on the JMTH dataset. ConVecs and SimDiffs were designed for semantic relation
classification, so it is not surprising that they perform much better
than balAPinc.

The KDSZ dataset was labeled using the substitutional definition of
lexical entailment (see Section~\ref{sec:defining}). On this dataset, there is
no statistically significant difference between any of the algorithms. This is
the ideal dataset for balAPinc, the dataset for which it was designed, so it is
natural that balAPinc has the highest accuracy. On the other hand, we see that the
two learning algorithms handle this dataset well, although they were
trained on the JMTH dataset (recall that this is the {\em different}
setup), which is quite different from the KDSZ dataset. It is good that
they are both able to cope with the qualitative difference between
the training data and the testing data.

All of the positive pairs in the BBDS dataset are instances of the
hyponym--hypernym semantic relation. Instances of this relation are
substitutable, so balAPinc is designed to handle them. ConVecs was
also designed specifically for this dataset, and we see from Table~\ref{tab:bbds-three}
that ConVecs reaches an accuracy of 87.6\% when the training data
is similar to the testing data. However, ConVecs has trouble bridging
the qualitative gap between the training data (the JMTH dataset) and
the testing data with the {\em different} setup. On the other hand,
SimDiffs is able to bridge this gap.

We have argued that the {\em different} evaluation is the most realistic
scenario, but it could be argued that the {\em entails} class is more
important than the {\em does not entail class}, and {\em entails}
is also more scarce in natural settings. Therefore Table~\ref{tab:summary2}
presents an alternative summary of the results. The table
reports ${\rm AP}_1$ instead of accuracy; this puts the emphasis
on the {\em entails} class. For the KDSZ and BBDS datasets, we report
the {\em clustered} setup. This is closer to the evaluation setup
of \citeA{kotlerman10}. In this table, we do not use bold font
to mark significant differences, because there is no agreement on
the appropriate statistical test for ${\rm AP}_1$.

\begin{table}
\caption{Summary of ${\rm AP}_1$ for the three algorithms with the three datasets}
\begin{minipage}{\textwidth}
\begin{tabular}{lccc}
\hline\hline
Algorithm  & JMTH ${\rm AP}_1$ & KDSZ ${\rm AP}_1$ & BBDS ${\rm AP}_1$ \\
\hline
balAPinc   &      0.56     & 0.37          &  0.73 \\
ConVecs    &      0.77     & 0.36          &  0.91 \\
SimDiffs   &      0.79     & 0.40          &  0.96  \\
\hline\hline
\end{tabular}
\end{minipage}
\label{tab:summary2}
\end{table}

Although Tables~\ref{tab:summary1} and \ref{tab:summary2} are based on
different scores and experimental setups, both support SimDiffs and the
similarity differences hypothesis. More generally, they suggest that second-order
features are useful for modeling lexical entailment. They also suggest that it
is beneficial to use two different spaces when constructing features for lexical
entailment.

Manually designing an asymmetric similarity measure is a difficult
task, as we can see from the equations in Section~\ref{subsec:balapinc}.
We believe that lexical entailment is more tractable when it is approached
as a supervised learning problem. The effort involved in manually designing
feature vectors is less than that required for designing similarity measures.
The performance of SimDiffs indicates that supervised learning can
yield better results than manually designing measures.

\section{Limitations and future work}
\label{sec:limits}

We have evaluated RLE directly, but most applications would use RLE
as a module inside a larger system. Future work will be needed to demonstrate
that our results with a direct evaluation can predict how an RLE module
will perform as a component of a larger system.

Although SimDiffs performs better than the competition, there is much
room for improved performance. However, when SimDiffs is used as a component
in a larger RTE system, words will be given in the contexts of sentences.
With the support of this contextual information and help from the
other modules in the system, SimDiffs might yield substantial
improvements in RTE performance. Related to this proposed future work,
\citeA{shnarch09} evaluated {\em lexical reference} rules \cite{glickman06}
derived from Wikipedia on the RTE-4 dataset. Used as a component in an RTE
system, the rules improved the RTE-4 score by 1\%.

Most of the past work on RLE has been based on the context inclusion hypothesis,
but ConVecs and SimDiffs show that other approaches, based on novel hypotheses,
can achieve competitive results. We believe that progress on the problem
will come from exploring a wide range of hypotheses and approaches. It is too
early at this stage of research to commit the field to a single hypothesis.

Recall the semantic relation subcategories hypothesis: Lexical entailment is not
a superset of high-level categories of semantic relations, but it is a superset of
lower-level subcategories of semantic relations. The experiments lend some support
to this hypothesis, but more research is needed. Any counterexamples for the hypothesis
could be handled by revising the taxonomy. However, if the required revisions become
onerous, then the hypothesis should be rejected.

The three algorithms here are based on three different hypotheses, but
all three achieve some degree of success on the task of RLE. This suggests
that it would be fruitful to combine the three approaches. One simple way
to combine them would be to average their real-valued outputs or apply voting
to their binary-valued outputs. This could be a useful direction for
future research.

We have focused here on individual words, but the natural next step is
to extend these ideas to phrases. \citeA{baroni12} have achieved promising
results with quantifier phrases, such as {\em all dogs} $\models$ {\em some dogs}.

Looking at Tables~\ref{tab:bejar-relation-categories-1-5}
and \ref{tab:bejar-relation-categories-6-10} in Section~\ref{subsec:jmth},
we see a high density of 1's ({\em entails}) for class-inclusion and
part-whole. The strong connection between these two categories and lexical
entailment may explain why \citeA{morris04} call hypernymy and meronymy
{\em classical relations}, whereas the relation in {\em chapel:funeral}
(space-time, location:activity, ID 9c) is {\em non-classical} (this is one
of their examples of a {\em non-classical} relation). For instance,
WordNet contains information about hypernymy and meronymy, but not
space-time relations. Particular relations might be considered {\em classical}
because we find them particularly useful for making inferences. This connection
is another topic for future work.

\section{Conclusion}
\label{sec:conclusion}

In this paper, we have evaluated three different algorithms for RLE on three
different datasets. Each algorithm relies on a different hypothesis about
lexical entailment. We find that SimDiffs has the best performance on two of
the three datasets. On the third dataset, there is no significant difference
in the three algorithms. The performance of SimDiffs suggests that similarity
differences make useful features for learning to recognize lexical entailment.

We have approached lexical entailment as a supervised learning problem of semantic
relation classification. The results indicate that this is
a promising approach to lexical entailment. This builds a bridge between
research in lexical entailment and research in semantic relation
classification. We hope that this connection will strengthen research in both fields.

\section*{Acknowledgements}

Thanks to Lili Kotlerman, Ido Dagan, Idan Szpektor, and Maayan Zhitomirsky-Geffet
for providing a copy of the KDSZ dataset and answering questions.
Thanks to Marco Baroni, Raffaella Bernardi, Ngoc-Quynh Do, and Chung-chieh Shan for
providing the BBDS dataset and answering questions. Thanks to the reviewers of
{\em Natural Language Engineering} for their very helpful comments.

%
%
\bibliography{lexical-entailment}
\bibliographystyle{theapa-modified}

\label{lastpage}

\end{document}